\documentclass{article}

\usepackage{arxiv}

\usepackage[utf8]{inputenc} 
\usepackage[T1]{fontenc}    
\usepackage{hyperref}       
\usepackage{url}            
\usepackage{booktabs}       
\usepackage{amsfonts}       
\usepackage{nicefrac}       
\usepackage{microtype}      
\usepackage{lipsum}		
\usepackage{graphicx}
\usepackage{natbib}
\usepackage{doi}
\usepackage{algorithm}
\usepackage{multirow, array}
\usepackage{algpseudocode}
\usepackage{amsmath}

\title{XAI meets Biology: A Comprehensive Review of Explainable AI in Bioinformatics Applications}

\author{ Zhongliang Zhou \\
	University of Georgia\\
	\And
	Mengxuan Hu \\
	University of Virginia \\
	\And
	Mariah Salcedo \\
	University of Georgia \\
	\And
	Nathan Gravel \\
	University of Georgia \\
	\AND
	Wayland Yeung \\
	University of Georgia \\
	\And
	Aarya Venkat \\
	University of Georgia \\
	\And
	Dongliang Guo \\
	University of Virginia \\
	\And
	Jielu Zhang \\
	University of Georgia \\
 	\And
	Natarajan Kannan\textsuperscript{*} \\
	University of Georgia \\
 	\And
	Sheng Li\thanks{corresponding author.} \\
	University of Virginia \\
}

\date{}

\renewcommand{\headeright}{}
\renewcommand{\undertitle}{}
\renewcommand{\shorttitle}{XAI meets Biology: A Comprehensive Review of Explainable AI in Bioinformatics Applications}

\hypersetup{
pdftitle={A template for the arxiv style},
pdfsubject={q-bio.NC, q-bio.QM},
pdfauthor={David S.~Hippocampus, Elias D.~Striatum},
pdfkeywords={First keyword, Second keyword, More},
}

\begin{document}
\maketitle

\begin{abstract}
Artificial intelligence (AI), particularly machine learning and deep learning models, has significantly impacted bioinformatics research by offering powerful tools for analyzing complex biological data. However, the lack of interpretability and transparency of these models presents challenges in leveraging these models for deeper biological insights and for generating testable hypotheses. Explainable AI (XAI) has emerged as a promising solution to enhance the transparency and interpretability of AI models in bioinformatics. This review provides a comprehensive analysis of various XAI techniques and their applications across various bioinformatics domains including DNA, RNA, and protein sequence analysis, structural analysis, gene expression and genome analysis, and bioimaging analysis. We introduce the most pertinent machine learning and XAI methods, then discuss their diverse applications and address the current limitations of available XAI tools. By offering insights into XAI's potential and challenges, this review aims to facilitate its practical implementation in bioinformatics research and help researchers navigate the landscape of XAI tools.
\end{abstract}



\section{Introduction}

Artificial intelligence is rapidly evolving, with critical advances made from deep learning and machine learning models \cite{lecun2015deep} in the fields of image \cite{krizhevsky2017imagenet} and speech recognition \cite{zhang2018deep}, natural language processing \cite{vaswani2017attention}, and robotics \cite{pierson2017deep}. Machine learning and deep learning models have been effectively integrated into different bioinformatics areas, including the biological sequence \cite{yan2022attentionsplice,zhou2023phosformer} and structure analysis \cite{taujale2021mapping,abbasi2020deepcda}, genome data analysis \cite{mieth2021deepcombi,chereda2021explaining}, bioimaging analysis \cite{mohagheghi2022developing,ukwuoma2022hybrid}. These models extract key information from biomedical databases to diagnose diseases \cite{mei2021machine}, identify drug targets \cite{bagherian2021machine}, and predict protein structures \cite{jumper2021highly}.  

Despite the many benefits of these AI models, they are limited by a lack of interpretability and transparency, commonly referred to as a ``black-box''. This black box represents an intrinsic disconnect between the model having learned underlying patterns from a dataset not apparent to human experts. Even if these patterns result in high-accuracy predictions, their obfuscation is a source of distrust, due to an inability to understand and validate the output, as well as delineate the limitations and biases of the model.

To pry open this black box, scientists have developed several promising Explainable AI (XAI) approaches to improve the transparency and interpretation of machine learning and deep learning models. In bioinformatics, XAI could help researchers identify the critical features that are most important for making accurate predictions and finding potential errors or biases in the models. This would enable researchers to improve the models and increase their accuracy and reliability.

In this review article, we present a thorough analysis of the Explainable AI (XAI) techniques and their applications in different bioinformatics research fields. Our objective is to deliver extensive guidance on the most effective use of XAI within a specific domain while promoting adherence to best practices. We begin with a brief introduction to the most relevant AI and XAI methods. Then, we discuss the use of AI in four different bioinformatics fields, and examine how XAI provides interpretability in different problem settings. Finally, we discuss the current limitations of the available XAI tools and identify areas for improvement in the context of bioinformatics research.

\section{Deep Learning Models for Bioinformatics}

AI models are generally classified into two categories: conventional machine learning (ML) models and deep learning (DL) models. Traditional ML algorithms require explicit human intervention for model correction and improvement. This iterative process enhances model interpretability but can become costly and impractical for solving increasingly complex problems. On the other hand, deep learning models leverage neural networks instead of human expertise. They excel in handling high-dimensional data and can automatically learn from raw input data. However, this comes at the cost of lower interpretability. Despite these trade-offs, deep learning models have gained significant traction in recent years thanks to its superior performance. Here, we focus on deep learning models, and we will briefly introduce three popular deep learning model architectures including Convolutional Neural Networks, Recurrent Neural Networks, and Attention Neural Networks.

\subsection{Deep Learning Models}
\paragraph{\textbf{Convolutional Neural Networks}}

\begin{figure}[!h]
    \centering
    \vspace{-5mm}
    \includegraphics[width=0.5\textwidth]{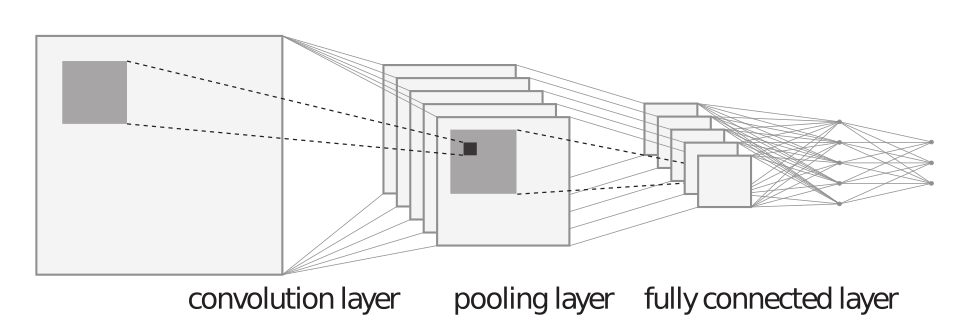}
    \caption{A demonstration of a classical CNN architecture}
    \vspace{-5mm}
    \label{fig:CNN}
\end{figure}

Typically, a Convolutional Neural Network consists of three types of components: convolution layers, pooling layers, and fully connected layers. Convolutional layers employ local filters to compute feature maps by sliding in both directions and using local patches as inputs. The shared filter weights across the entire frame reduce the number of trainable parameters. Pooling layers, typically either max pooling or average pooling, perform subsampling in the feature maps, allowing the model to aggregate local features and identify more complex features. The fully connected layer connects every neuron from the preceding layer to every neuron in the next layer. It helps combine the learned features from the convolutional and pooling layers and make predictions. Although CNNs were initially designed for handling 2D data such as images, they have been shown to manage biological sequential data effectively \cite{kim-2014-convolutional, taujale2021mapping}.

\paragraph{\textbf{Recurrent Neural Networks}}
\begin{figure}[!h]
    \centering
    \vspace{-5mm}
    \includegraphics[width=0.5\textwidth]{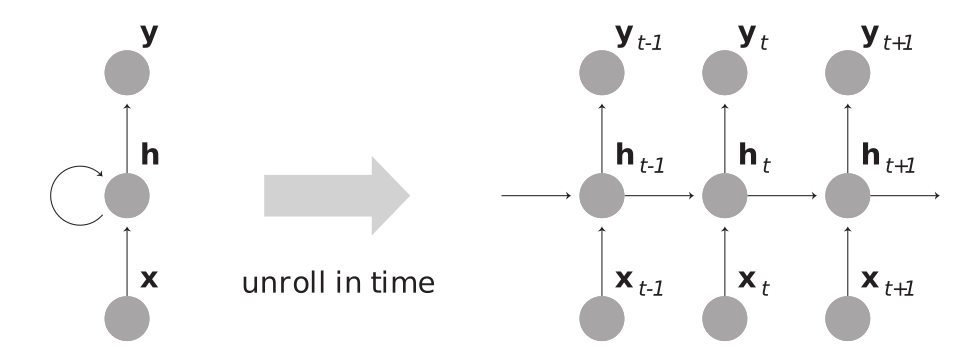}
    \caption{A demonstration of a classical RNN architecture}
    \vspace{-5mm}
    \label{fig:rnn}
\end{figure}
Recurrent Neural Networks (RNNs) are a widely used deep learning architecture featuring cyclic connections. Traditional RNNs process inputs one at a time and store previous information in hidden states $h$. When a new input arrives, the model considers both the current state $h_{t}$ and the previous hidden state $h_{t-1}$ to generate a new output. In cases where inputs from both the past and future are crucial, bidirectional recurrent neural networks (BRNNs) are designed and employed extensively. The RNN model can accommodate arbitrary input lengths and output a single hidden state as the final output $h_{t}$. This feature is particularly advantageous in bioinformatics since many DNA, RNA, protein, and single-cell transcriptomics data have variable lengths. 

\paragraph{\textbf{Attention Neural Networks}}
\begin{figure}[!h]
    \centering
    \vspace{-5mm}
    \includegraphics[width=0.5\textwidth]{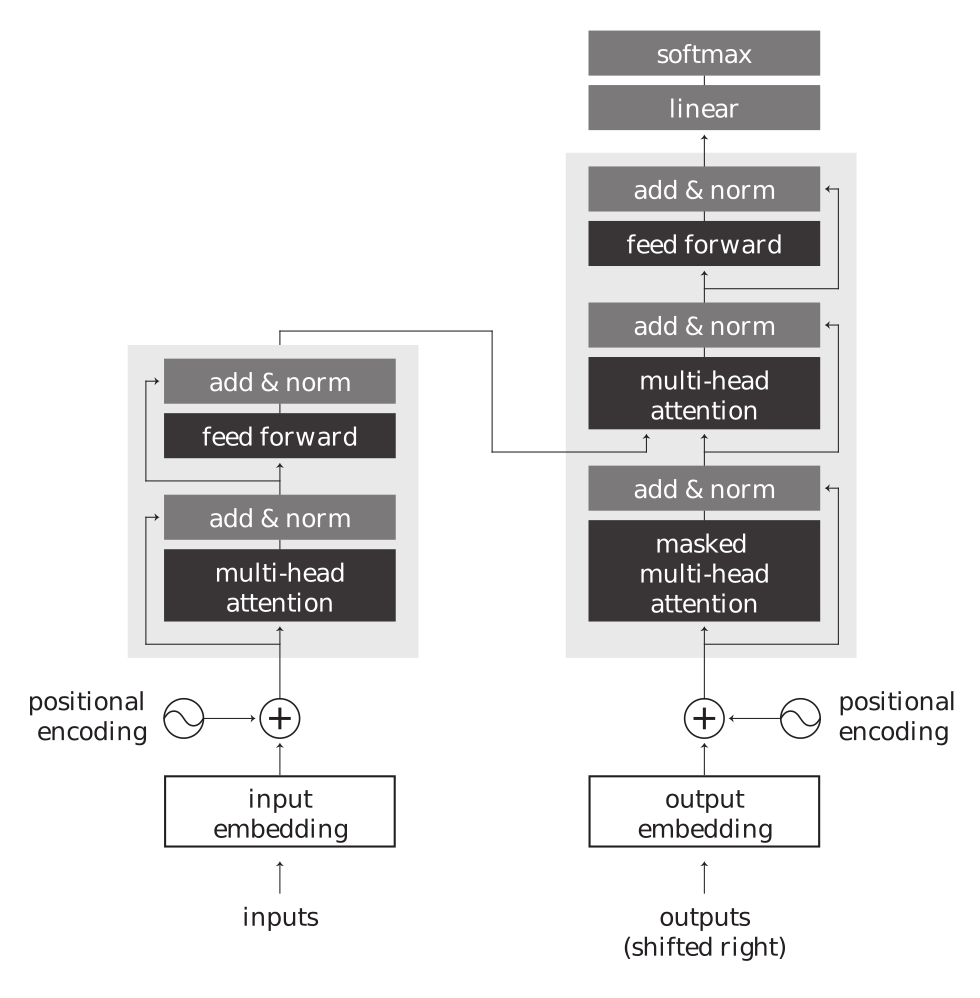}
    \vspace{-5mm}
    \caption{A demonstration of a classical Attention architecture}
    \vspace{-5mm}
    \label{fig:attention}
\end{figure}
The attention mechanism \cite{vaswani2017attention}, has emerged as a powerful technique for various applications in computer vision and natural language processing. By mimicking cognitive attention, the attention mechanism allows models to focus on the most relevant information while disregarding irrelevant input portions. This selective focus enables the model to capture long-range relationships more effectively, as it can directly compare positions regardless of their distance within the sequence. In the field of bioinformatics, models that utilized attention architecture have achieved remarkable success in tasks like protein folding prediction \cite{lin2022language}, functional classification \cite{zhou2023phosformer,yeung2023tree}, and mutation effect prediction \cite{yeung2023alignment}.

\section{XAI Methods for Bioinformatics}

\begin{table*}[!th]
    \centering
    \resizebox{\textwidth}{!}
    {\begin{tabular}{l|ll|l}
\hline
\textbf{\textbf{Category}} &
  \multicolumn{2}{l|}{\textbf{Methods}} &
  \textbf{\textbf{Articles}} \\ \hline
\multirow{3}{*}{Model-agnostic methods} &
  \multicolumn{2}{l|}{Local Interpretable Model-agnostic Explanations (LIME)} &
  LIME for Bioimage ( \cite{madhavi2022efficient}, \cite{magesh2020explainable}, \cite{hassan2022prostate}) \\ \cline{2-3}
 &
  \multicolumn{2}{l|}{SHapley Additive exPlanations (SHAP)} &
  \begin{tabular}[c]{@{}l@{}}SHAP for Bioimage(\cite{van2020volumetric}, \cite{foroushani2022accelerating}, \cite{van2020volumetric}),\\ SHAP for expression data (\cite{yap2021verifying}, \cite{yu2021explainable}, \cite{nelson2022smash}, \cite{kuruc2022stratified}, \cite{lemsara2020pathme},\cite{withnell2021xomivae}),\\ SHAP for biological sequences(\cite{yang2023cfa}),\\ SHAP for biological structure (\cite{gu2022protein})     \end{tabular} \\ \cline{2-4} 
 &
  \multicolumn{2}{l|}{Layer-Wise Relevance Propagation(LRP)} &
  LRP for expression data ( \cite{mieth2021deepcombi}, \cite{chereda2021explaining}) \\ \hline
\multirow{3}{*}{Model-specific methods} &
  \multicolumn{2}{l|}{Class Activation Maps(CAM)} &
  \begin{tabular}[c]{@{}l@{}}Grad-CAM for Bioimage( \cite{wang2020explainable}, \cite{altan2022deepoct}, \cite{singh2021covidscreen}, \cite{yang2022deep}, \cite{karim2020deepcovidexplainer}, \cite{nedumkunnel2021explainable}, \cite{ukwuoma2022hybrid},\cite{vasquez2022interactive}, \cite{islam2022explainable}, \cite{jahmunah2022explainable}),\\ Grad-CAM for expression data (  \cite{karim2019onconetexplainer}, \cite{lombardo2022deepclasspathway}),\\ Grad-CAM for biological sequence (\cite{chen2021xdeep}, \cite{monteiro2022explainable}), \\ Grad-CAM for biological structure (\cite{mukherjee2021deep},   \cite{taujale2021mapping},  \cite{yang2022mgraphdta})  \end{tabular} \\ \cline{2-4} 
 &
  \multicolumn{2}{l|}{Attention Scores} &
  \begin{tabular}[c]{@{}l@{}} Attention Scores for biological sequences(\cite{yan2022attentionsplice},  \cite{danilevicz2022dnabert},  \cite{avsec2021effective},  \cite{hu2019deephint},  \cite{tian2021deephpv},  \cite{liang2021deepebv},  \cite{mao2017modeling}), \\   Attention Scores for biological structures (\cite{abbasi2020deepcda},   \cite{karimi2020explainable},   \cite{karimi2019deepaffinity},   \cite{gao2018interpretable} )\end{tabular} \\ \cline{2-4} 
 &
  \multicolumn{2}{l|}{Self-Explainable Neural Network} &
  \begin{tabular}[c]{@{}l@{}}KP Neuron for expression data( \cite{bourgeais2021deep}, \cite{park2021classification}, \cite{hao2019interpretable}, \cite{fortelny2020knowledge},  \cite{liu2020fully},  \cite{gut2021pmvae}, \cite{rybakov2020learning}, \cite{lotfollahi2021learning})\end{tabular} \\ \hline
\end{tabular}}
\caption{Categorization of XAI methods in different bioinformatics applications based on their properties}
    \label{tab:xai_methods}

\end{table*}

We categorize XAI tools into two primary groups. The first group consists of model-agnostic methods, which can be applied to multiple ML or DL models, while the second group, model-specific methods, can only be applied to a particular type of ML or DL model. 

\subsection{Model-agnostic Methods}

\paragraph{\textbf{Local Interpretable Model-agnostic Explanations (LIME)}}
Local Interpretable Model-agnostic Explanations (LIME) \cite{ribeiro2016should} is a methodology for generating local explanations of ML models. Given a data instance $x$, LIME generates a set of perturbed instances $\{x_1, x_2, \ldots, x_n\}$ and obtains the predictions of the complex model on these instances. Then, LIME trains a linear model $g$ on these generated points while minimizing the following objective function:
\begin{equation}
\text{argmin}_g \sum_{i=1}^n w_i L(f(x_i), g(x_i)) + \Omega(g),
\end{equation}
where $f(x_i)$ is the prediction of the complex model on instance $x_i$, $g(x_i)$ is the prediction of the linear model on instance $x_i$, $L(f(x_i), g(x_i))$ is the loss function comparing the predictions of the complex model and the linear model, and $w_i$ is the weight assigned to instance $x_i$, which is a function of the distance between $x_i$ and $x$. The closer the instance $x_i$ is to $x$, the higher the weight $w_i$. $\Omega(g)$ is a regularization term that controls the complexity of the linear model. 

By minimizing this objective function, LIME finds an interpretable linear model that approximates the complex model's behavior locally around the instance $x$. The coefficients of the linear model can then be used to explain the prediction of the complex model for that instance.

\paragraph{\textbf{SHapley Additive exPlanations (SHAP)}} 

SHapley Additive exPlanations (SHAP) \cite{lundberg2017unified} is a method used to explain the outputs of ML models. It is based on the concept of Shapley values from cooperative game theory \cite{shapley1953value}, which allocates the contributions of each feature to the prediction for a specific instance. The method helps in understanding the impact of each feature on the prediction and can provide insights into the model's behavior.

Given a model $f$, the Shapley value $\phi_j$ for the $j$-th feature is defined as:
\begin{equation}
\phi_j(x) = \sum_{S \subseteq N \setminus \{j\}} \frac{|S|!(|N|-|S|-1)!}{|N|!} [f_{S \cup \{j\}}(x) - f_S(x)],
\end{equation}
where $N = {1, 2, \ldots, n}$ is the set of all features, and $|S|$ denotes the cardinality of the set $S$.
$|N|$ denotes the cardinality of the set $N$. $f_S \cup {j}(x)$ is the prediction of the model when the features in $S$ and the $j$-th feature are considered.
$f_S(x)$ is the prediction of the model when only the features in $S$ are considered.

The Shapley value $\phi_j(x)$ represents the average marginal contribution of the $j$-th feature across all possible combinations of features. A positive SHAP value for a feature indicates an increase in the prediction relative to the baseline, while a negative value signifies a decrease. The magnitude of a SHAP value represents the strength of the feature's impact on the prediction.

\paragraph{\textbf{Layer-Wise Relevance Propagation (LRP)}}

Layer-wise Relevance Propagation (LPR) is a method used to explain the predictions of deep neural networks by attributing relevance scores to the input features. While both IG and LRP aim to explain the importance of features or neurons in a neural network, IG focuses on the gradients and their interactions, whereas LRP is based on the backward propagation of the output through the layers of the network. LRP involves three steps: (1) Compute the relevance score for the output neuron corresponding to the predicted class. (2) Propagate the relevance scores backward through the network from the output layer to the input layer using the LRP formula. (3) Obtain the relevance scores for the input features. The LRP formula for step 2 is:
\begin{equation}
R_j = \sum_{k} \frac{a_j w_{jk}}{\sum_{j} a_j w_{jk}} R_k,
\end{equation}
where $R_j$ is the relevance score of neuron j in the current layer, $R_k$ is the relevance score of neuron k in the next layer of j, $a_j$ is the activation of neuron j in the current layer, $w_jk$ is the weight connecting neuron j in the current layer to neuron k in the next layer, $\sum_k$ denotes the summation over all neurons k in the next layer, and $\sum_j$ denotes the summation over all neurons j in the current layer. To calculate the relevance score for each neuron, initialize the relevance score for the output neuron corresponding to the predicted class, and propagate the relevance scores backward using the LRP formula for each layer until reaching the input layer. The final relevance scores for the input features will be the relevance scores $R_i$ for each input neuron $i$.

\subsection{Model-specific Methods}

\paragraph{\textbf{Class Activation Maps (CAM)}}
\begin{figure}[!t]
    \centering
    \includegraphics[width=0.5\columnwidth]{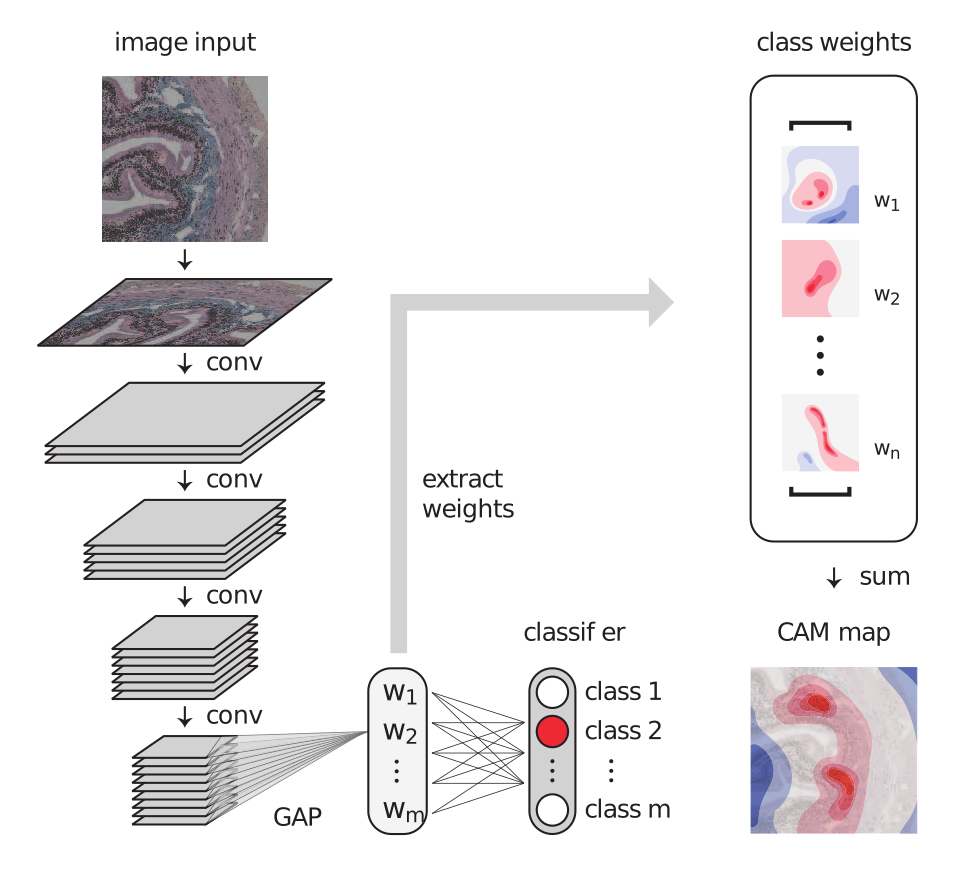}
    \caption{Procedure of CAM. In the final CAM map, The more vibrant the color, the more significant it represents.}
    \label{fig:CAM}
    \vspace{-5mm}
\end{figure}

Class Activation Maps is a classical XAI method that essentially generates a heatmap that indicates the portion of the image responsible for CNN prediction by adopting global average pooling (GAP) \cite{zhou2016learning}. GAP computes the average value for each feature map in the last convolutional layer. For example, for a $k$-th feature map of the last layer with width $m$ and height $n$, the average value is the average of all activation values $f_k(x,y)$ in the $k$-th feature map:
\begin{equation}
    w_k=\frac{1}{mn}\sum_{x}^m\sum_{y}^nf_k(x,y)
    \label{equ: weight}
\end{equation}
Those average values $w={w_1, w_2,..., w_K}$ ($K$ is the total number of feature maps in the last convolutional layer) are then used as the features of the last fully connected layer to predict the output. The weight $w_k^c$ of the average value $w_k$ of the corresponding feature map $f_k$ with respect to class $c$ in the above prediction is adopted for the final heatmap calculation denoted as CAM. To get CAM for a specific class, $CAM_c$ is obtained by adding the weighted feature maps in the last convolution layer. The whole process is shown in Figure \ref{fig:CAM}. In addition to the original CAM method, later modification\cite{selvaraju2017grad} remove the need of specialized model architecture thus can be adapted to explain more tasks and models \cite{taujale2021mapping}.

\paragraph{\textbf{Attention Scores}}

\begin{figure}[!b]
    \centering
    \includegraphics[width=0.5\columnwidth]{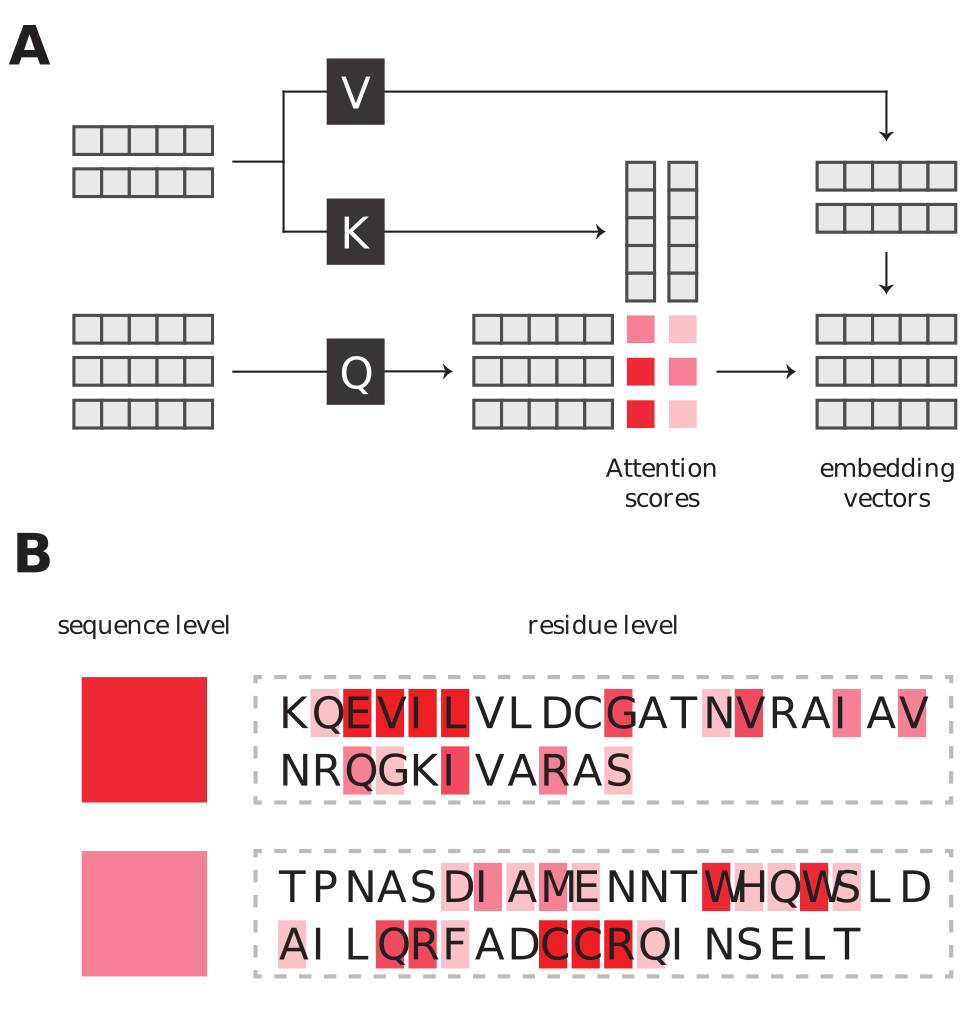}
    \caption{Example of attention scores.}
    \label{fig:Attention}
    \vspace{-5mm}
\end{figure}

In attention-based models \cite{vaswani2017attention}, attention scores are calculated to determine the importance of different input elements for a specific output element. To calculate attention scores, input is first converted into embeddings and then projected into Query ($Q$), Key ($K$), and Value ($V$) vectors using three learnable weight matrices ($W_Q$, $W_K$, and $W_V$). Attention scores are computed as the dot product between the Query ($Q_i$) and Key ($K_j$) vectors, normalized by the square root of their dimension ($d_k$). For a given position, the Query would be itself while the Keys and Values would be all other positions. The softmax function is applied to these scores to obtain normalized probabilities, which are then multiplied with their respective Value ($V_j$) vectors and summed to generate the output. In practice, this process is performed in parallel for multiple attention heads, capturing different types of relationships, and the results are concatenated and projected through a linear layer to produce the final output. We can then visualize the attention scores to identify which input tokens had the highest scores and thus the highest importance in the model's decision-making process as shown in Figure \ref{fig:Attention}.

\section{Applications}

In bioinformatics, machine learning techniques have been employed to make predictions and employ explainable artificial intelligence (XAI) tools to elucidate these predictions. In this review, we classify these applications into four principal categories: sequence analysis, structural analysis, genome analysis, and bioimaging analysis. In subsequent sections, we will begin with a comprehensive introduction to the overall applications. Subsequently, we will delve into various XAI tools, distinguishing between model-agnostic and model-specific methods, and we will discuss how each XAI tool contributes to uncovering the insights derived from the constructed machine learning models in the context of bioinformatics.

\subsection{Biological Sequence}

\begin{figure*}[!th]
\centering
\includegraphics[width=\textwidth]{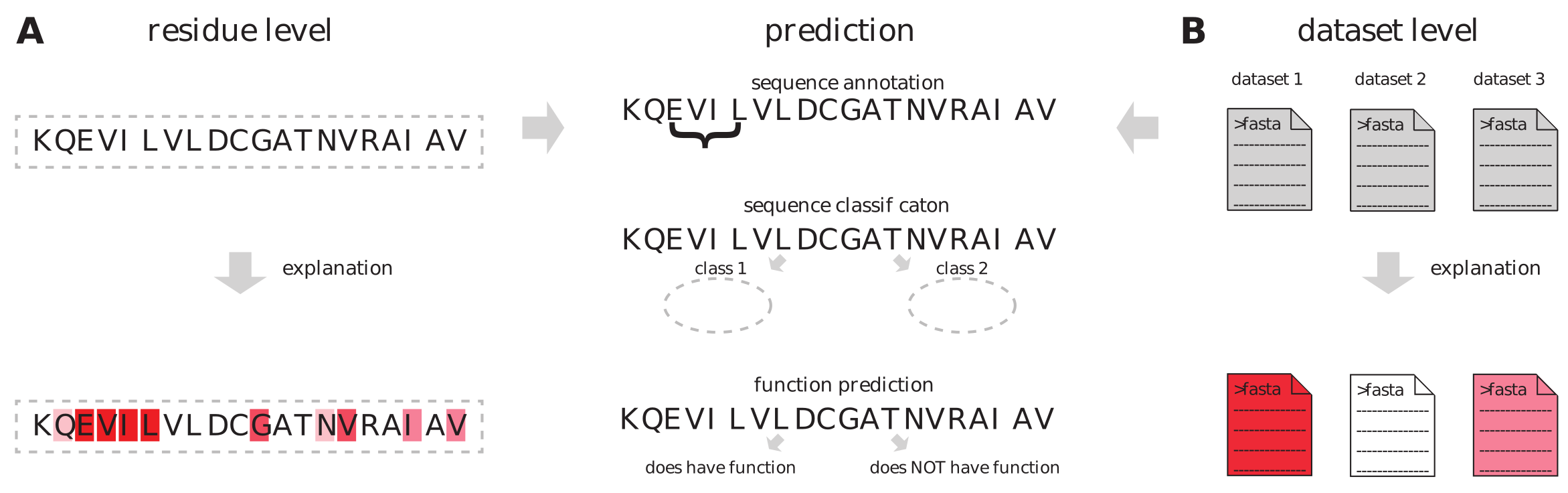}
\vspace{-5mm}

\caption{Sequence base XAI methods primarily focus on highlighting the input of the DL models. Methods such as SHAP, attention scores, Grad-CAM, and other methods will highlight (A) regions of the sequences on a character or k-mer level of individual sequences. (B) Other more broad XAI approaches will highlight whole data sets or subsets of the training data. The features highlighted will give clues to DL models' reasoning behind the downs stream predictive task of the model.}
\label{fig:sequence}
\vspace{-5mm}
\end{figure*}

Sequence-based ML models have shown promising results in the field of biology to take discrete token-based data such as DNA, RNA, or protein sequences in order to predict biologically important tasks like protein classification, functional association, disease diagnostics, regulatory association, etc. However, due to the complexity of the recent popular DL models, it is challenging to understand how a DL model may have arrived to a particular conclusion about the predictions. To bridge this gap in knowledge, many DL models employ XAI methods such as attention, Grad-CAM, or SHAP to uncover underlying features contributing to the prediction. The following sequence-based XAI methods illuminate the ``Black Box" dilemma and reveal a deeper understanding, from the DNN models, of the biological theory behind the initial predictive task without specific prior training.
  
\subsubsection{Model-Agnostic Methods}
 
\paragraph{\textbf{SHAP}}
Transcriptional regulation is controlled by cis-regulatory modules (CRMs) that are widespread in eukaryotic genomes. Epigentic factors such as histone binding provide additional layers of transcriptional control by CRMs \cite{schoenfelder2019long,berman2002exploiting}. The CRM Function Annotator (CFA) model \cite{yang2023cfa}, a CNN model tasked with categorizing CRMs as enhancers, promoters, or insulators, utilizes the SHAP method to investigate the participation of different training sets on the overall performance of the model. Each training set was inherently connected to a unique histone due to the nature of the ChIP-seq training data sets \cite{yang2023cfa}. SHAP values were assigned to histone-associated training data sets that match with a previously known binding preference of the histone subunits. The authors demonstrated that the CFA model learned this histone binding preference without being explicitly trained on the task in order to achieve the original prediction task of CRM categorization. 

\subsubsection{Model Specific Methods}
\paragraph{\textbf{Class Activation Maps}}
xDeep-AcPEP \cite{chen2021xdeep} is a CNN-based model tasked with finding novel Anticancer peptides (ACPs), a promising alternative therapeutic agent for treating cancers. The model utilizes peptide sequences as input and predicts the lethality of the input peptide against six cancer cell types. Grad-CAM is then used to validate the contribution of the peptide on a per-residue basis. Of the per-residue contribution scoring, a distribution of all the residues was plotted by cancer types against the residue’s average importance score. This revealed high Grad-CAM values for hydrophobic residues in the peptide sequence that matched previously reported amino acid composition in CancerPPD \cite{tyagi2015cancerppd} data set. This would suggest that the model learned deeper biochemical constraints of each residue type without being explicitly trained. Another CNN-based model that utilizes Grad-CAM was proposed by the Moteiro et al. \cite{monteiro2022explainable} to explain predictions of the binding affinity of proteins and small molecules. Grad-CAM aided in creating a global average pooled weights, resulting in neuron importance weights that capture the contribution of regions of the input sequences. The scores were then mapped back to essential motifs previously known to aid in binding and compound protein crystal structures \cite{desaphy2015sc}. As an XAI method, Grad-CAM can uncover a deeper understanding of how these models produce predictions.

\paragraph{\textbf{Attention Scores}}

The attention-based model often offers innate explainability by the process of calculating the attention matrix. In the Phosformer model \cite{zhou2023phosformer} for kinase-specific phosphorylation prediction, the attention values are utilized to validate the model's ability to recognize protein substrate specificity motifs by focusing on specific substrate residues. 
Through the final Attention layer of the model's encoder, it was found that Phosformer learns to identify substrate specificity determinants post-fine-tuning. The model directed attention to specific positions in the peptide substrate that are biochemically known to contribute to kinase substrate specificity. 

In addition, Attention-based neural networks have been extensively employed to characterize Cis-regulatory modules (CRMs) and their role in transcription regulation. By incorporating an attention layer, performance is enhanced via training, allowing us to identify the model's prioritized features for decision-making.

AttentionSplice \cite{yan2022attentionsplice} predicts splice acceptor and donor sites from protein-coding gene DNA sequences, offering insight into alternative splicing mechanisms. This process allows additional regulation and specialization opportunities for eukaryotic genes \cite{breathnach1981organization}. AttentionSplice employs the attention layer to assign importance to each nucleotide in the input sequence. This approach was validated by successfully identifying known human DNA transcription motifs from the HOCOMOCO database \cite{kulakovskiy2013hocomoco}, without explicit training. 

Other models detecting functional genomic features include Danilevicz et al.'s model \cite{danilevicz2022dnabert}, which identifies lncRNAs, and the Enformer model \cite{avsec2021effective}, which predicts gene expression and chromatin states in humans and mice. Both models, like AttentionSplice, utilize attention scores for XAI validation. Furthermore, DNNs can assist in disease diagnostics. DeeP-HINT \cite{hu2019deephint}, a CNN with an attention layer, can detect HIV integration sites in the human genome. Other diagnostic models, such as DeepHPV \cite{tian2021deephpv} and DeepEBV \cite{liang2021deepebv}, focus on Human Papillomavirus and Epstein-Barr virus detection, respectively. These models employ XAI methods, aligning attention scores with known transcription factor binding sites or related oncogenic virus integration sites.

Lastly, EPIANN \cite{mao2017modeling} successfully predicted promoter and enhancer segments from DNA sequences, using attention scores to understand the model's rationale. Each mentioned model utilized an attention score for validation, demonstrating the potential for DNNs to learn independent biology-related tasks and aid in predictions without explicit prompting.

\subsubsection{Other}

A CNN model \cite{liu2022interpretation} developed by  Liu et al. was tasked with predicting fiber production on Upland cotton plants (Gossypium hirsutum) using genetic data as input. To increase confidence in the model's prediction of fiber production, a position weight matrix was generated from the DeepLIFT XAI method and highlighted sequence motifs that had the highest impact on the predictions. The DNA segments highlighted were then compared to already known motifs of transcriptional factors linked with fiber production which were previously characterized and cataloged in the JASPER \cite{khan2018jaspar} database. Their results suggest that the model learned to identify crucial motifs associated with fiber production independent of the original phenotype prediction training objective. 

\subsection{Biological Structure}

\begin{figure*}[!th]
\centering
\includegraphics[width=\textwidth]{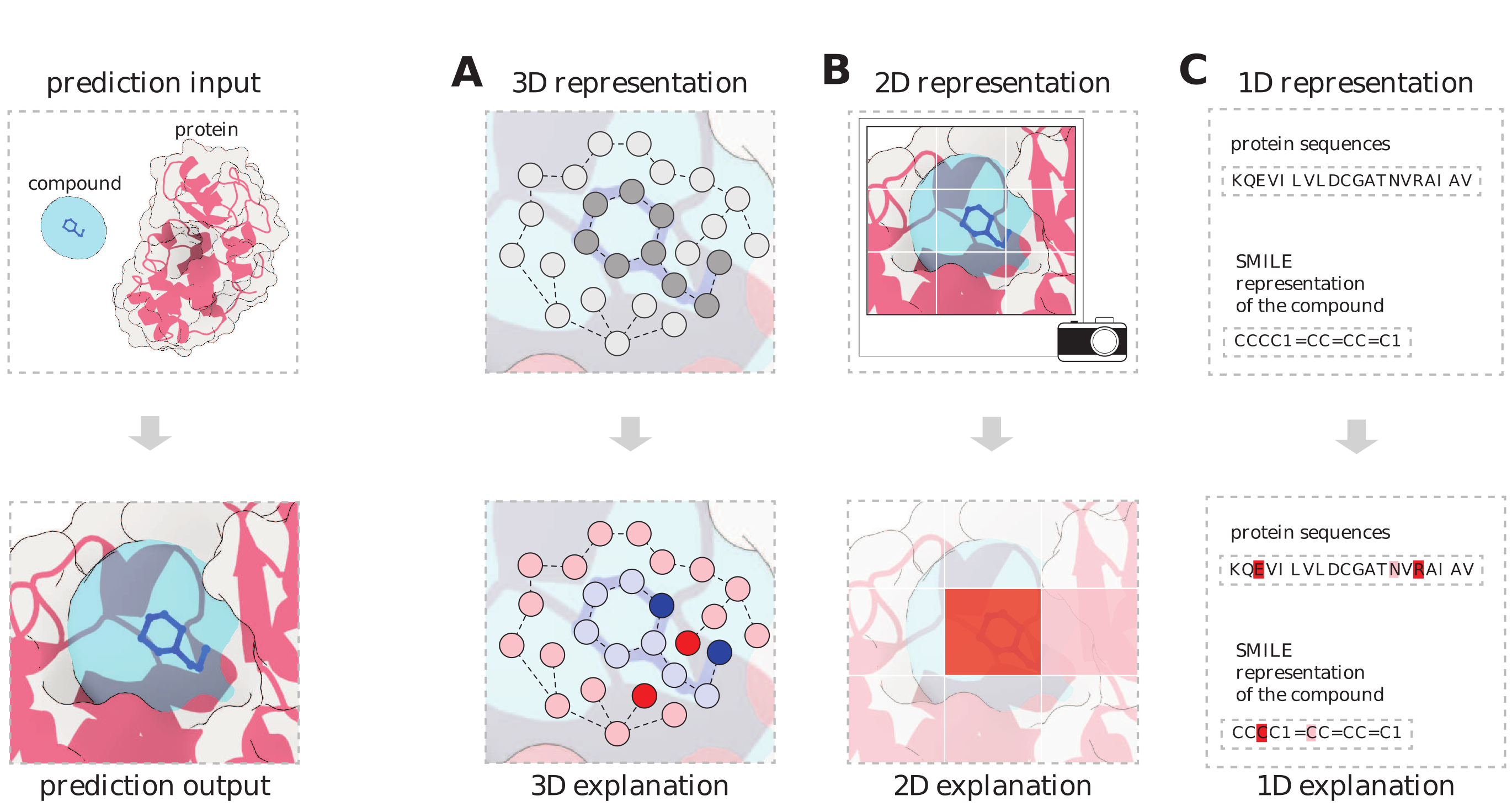}
\vspace{-5mm}
\caption{XAI methods cater to various data types used as inputs for DL models predicting biological structures by highlighting relevant input features differently. (A) Graph representations of proteins and small molecules depict each atom in 3D space, capturing the 3D context required for binding. (B) 2D images of proteins and small molecules are divided into sections, with important regions highlighted for their contribution to the binding. (C) Sequence representations of proteins and small molecules in 1D space, focusing on essential characters contributing to the binding.}
\label{fig:knowledge_graph}
\vspace{-5mm}
\end{figure*}

Biological structures like proteins and small molecules can be represented in various dimensions, including 1D sequences, 2D pictures or graphs, and 3D coordinates. Structure-based DNN models primarily focus on the prediction of drug-protein or protein-protein interactions, prediction of protein structures, and prediction of small molecule function. Predicting drug-protein interactions is vital for drug discovery \cite{abbasi2020deepcda}. Although DL has shown great promise in results, the lack of XAI in DL can leave some predictions unusable in certain fields. For practical applications of DNNs in health care, a certain level of transparency from the models is expected. XAI such as SHAP, LIME, Grad-CAM, and, attentions scores can help alleviate such concerns. 

\subsubsection{Model-Agnostic Methods} 

\paragraph{\textbf{SHAP}}

 A model that highlights SHAP as a structure-based XAI method is SEGSA-DTA \cite{gu2022protein}. This graph CNN represents the input of the protein binding pocket and small molecule as a connected graph and predicts binding affinity between drugs and proteins. SHAP as an XAI method was implemented on proteins and small molecule pairs with known binding properties \cite{kurumbail1996structural,wang2013structural} and tested on mutant proteins where the binding residues are mutated \cite{luo2017allosteric,rodriguez2014structure,ehrlich2017towards}. The SHAP importance values could detect the mutations that affect the binding of the small molecule and additionally could highlight the key residues that play a role in binding to the small molecule of interest. Interestingly the areas highlighted were all in the binding interface- a pattern that matches human intuition for understanding compound-protein interactions. 


\subsubsection{Model Specific Methods}
\paragraph{\textbf{Class Activation Maps}}
A typical example of using Class Activation Maps in sequence structure data is the GT-CNN \cite{taujale2021mapping} model that was developed to predict glycosyltransferase fold and family types based on secondary structure annotation. The GT-CNN model was further analyzed using the Grad-CAM method. Specifically, mapping the CAM values to the input sequence revealed high activation values for conserved regions in the GT-A core, such as the DXD motif, G-loop, and the first two beta-sheets of the characteristic Rossmann fold that are known to structurally differentiate GT-A fold from other fold types. Thus the Grad-CAM method enabled structural interpretation of the prediction without directly involving structural data in the training.   

In small molecule function prediction, a modified CNN model predicts the capacity of SMILE representations of small molecules as input and determines endocrine-disrupting potential as output \cite{mukherjee2021deep} With endocrine-disrupting molecules being linked with toxicity in humans, chemists have classified functional groups that assist in dangerous compounds including endocrine-disruptors with the metrics such as "critical structural motifs" \cite{mukherjee2021deep} and "structural alerts" \cite{alves2016alarms}. XAI is achieved through Grad-CAM, highlighting structural motifs and comparing them to known "critical structural motifs" and ``structural alerts'' related to the molecules' overall toxicity \cite{casals2011endocrine}. The model learned to distinguish these motifs without explicit toxicity detection training. Additionally, a GRU RNN Bung et al. model \cite{bung2022silico} predicting compound-protein interactions employed Grad-CAM to identify binding interfaces without explicit training.

Comparing XAI methods is crucial to find the best approach for different models and tasks. The GNN model MGraphDTA \cite{yang2022mgraphdta} compared class activation mapping and attention-based XAI methods for compound-protein interaction prediction. The model was designed to test Grad-AAM on both fully convolutional and attention-layer variants. All three methods could assign atom-specific importance values, with the benchmark being alignment with "structural alerts" \cite{alves2016alarms} labels from the ToxCast database \cite{kavlock2012update}. The Grad-AAM method on the fully convolutional variant had the highest accuracy and precision, highlighting the importance of choosing the appropriate XAI method for a given model and task.

\paragraph{\textbf{Attention Scores}}

Attention-based methods can also be applied to protein structure analysis. Deep Cross-Domain compound-protein Affinity (DeepCDA) \cite{abbasi2020deepcda} is a hybrid model employing convolutional and long-short-term memory layers with an attention mechanism to predict compound-protein interactions. The attention layer serves as an XAI method, highlighting input sequence regions involved in binding for both protein and compound SMILE representations. The attention values, encoding binding strength, were compared to known crystal structures, revealing that the model learned key residues in the binding interface without explicit training.

Comparing XAI methods across different DNN models, such as the hierarchical RNN model DeepRelations \cite{karimi2020explainable}, DeepAffinity+ \cite{karimi2019deepaffinity}, and Gao et al. \cite{gao2018interpretable}, can provide valuable benchmarks for validation and understanding of the model's inner workings. All three models used attention-based XAI to highlight important features in protein structures and small molecules, independently identifying similar key residues and atoms, demonstrating the potential for additional validation when exploring XAI.

\subsection{Gene Expression and Genomics Data} 

Transcriptomics, genomics, and epigenomic methods reveal complex biological systems regulated by numerous gene interactions. DL models have shown promise in processing sequencing data, patient clustering, and disease classification. However, interpreting and validating DL-model outputs can be challenging due to model architecture (CNN and autoencoders) and data complexity. Explainable AI (XAI) methods can identify critical genes/pathways, providing biological context for predictions. This section explores XAI approaches for understanding and modeling these data types.

\subsubsection{Model Agnostic methods} 

\paragraph{\textbf{Layer-wise relevance propagation (LRP)}} 

DeepCOMBI uses LRP to explore gene and phenotype relationships using data from Genome-wide association studies (GWAS)\cite{mieth2021deepcombi}. GWAS primarily seeks to investigate phenotypic effects of single nucleotide polymorphisms (SNPs).  LRP creates importance scores highlighting important SNPs for the phenotypes a neural network assigns to each subject, thus providing relevant locations for multiple hypothesis testing.  Using this method, the authors were able to identify two novel disease associations (for hypertension and diabetes). 

Chereda et al.\cite{chereda2021explaining} proposed the use of similar methods for graph-CNNs known as Graph Layer-wise Relevance Propagation (GLRP). By assigning gene expression data as vertices in the graph, the authors predicted metastatic events in breast cancer, and using GLRP they were able to create patient-specific molecular subnetworks that largely agree with clinical knowledge and identify common and novel drivers of tumor progression.     

\subsubsection{Model Specific methods} 

\begin{figure*}[!th]
\centering
\includegraphics[width=\textwidth]{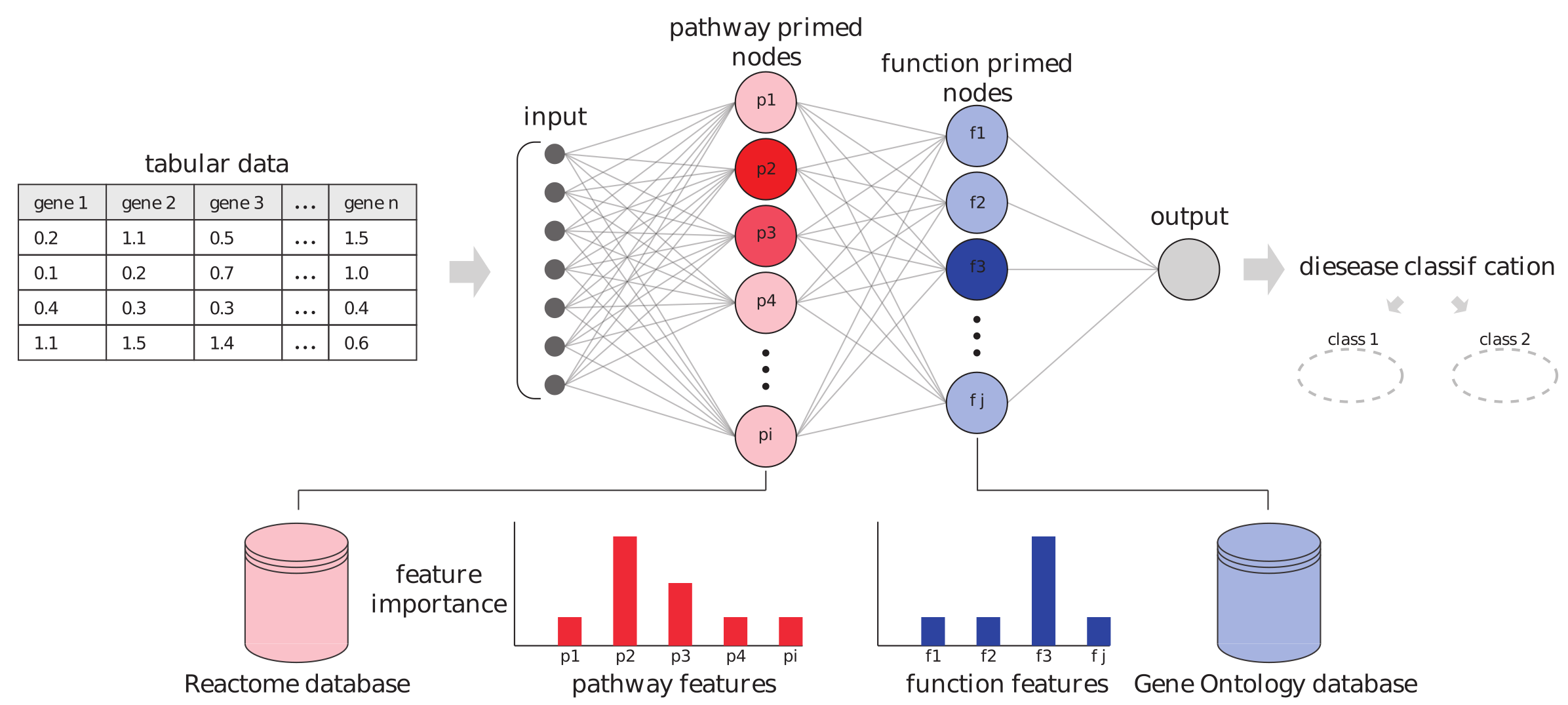}
\vspace{-5mm}
\caption{Self-explainable XAI method in which tabular expression data is used for a classification task. Through the use of knowledge-primed CNNs, the feature importance of each node can be assessed and interpreted for a particular output. }
\label{fig:knowledge_graph}

\vspace{-5mm}
\end{figure*} 

\paragraph{\textbf{SHAP}} 

XAE4Exp (eXplainable AutoEncoder for Expression data), integrates the use of SHAP with expression data, which allows for quantitative evaluation of each gene’s contribution to the hidden layers of an auto encoder\cite{yu2021explainable}.  This method was used to highlight the nonlinear contribution of genes that contribute to breast cancer but are not differentially expressed. SHAP methods have additionally been utilized for tissue classification with RNA-seq data \cite{yap2021verifying} and cell-type classification with single-cell sequencing (scRNA-seq) data as demonstrated by the user-friendly SMaSH\cite{nelson2022smash}.  

DeepSHAP has been used to examine how the changes in loss of function affect the genes deemed most important when combining RNA-seq data from different sources \cite{kuruc2022stratified}. In this study, the authors seek to address a major issue with utilizing DL models for identifying biomarkers in clinical studies, in which data is typically combined due to sparsity-meaning loss functions that assume some level of data uniformity is inappropriate to use. By using Deep SHAP and comparing it with known biomarkers, Kuruc et al. were able to prove the validity of their use of partial likelihood as a loss function.   

Deep SHAP has also been applied to multi-omics data.  Multi-omics data analysis encompasses capturing variations at multiple levels including the genome, epigenome, transcriptome, proteome, and metabolome. Properly integrating and analyzing this data can be a challenge.  To address these challenges, tools such as PathME \cite{lemsara2020pathme} have been developed.  

Pathme is a multi-modal sparse auto-encoder tasked with clustering multi-omic patient data that uses Deep SHAP to identify which genes were most influential for each patient cluster for a variety of cancer datasets which consisted of gene expression, miRNA expression, DNA methylation and CNVs. XOmiVAE (eXplainable “omics” variational autoencoder) similarly uses Deep SHAP for highlighting features used for clustering patients according to cancer classification using gene expression and DNA methylation data as input \cite{withnell2021xomivae}.   

\paragraph{\textbf{Grad-CAM}} 
OncoNetExplainer, a model trained on a large pan-cancer dataset tasked with cancer classification utilized Grad-CAM++ to create a ranking of genes as features and create class-specific heat maps identifying relevant biomarkers \cite{karim2019onconetexplainer}. 

Grad-CAM saliency maps can also visualize feature-importance for cancer classification tasks using transcriptomic data as demonstrated by the method proposed by Lombardo that predicts HPV-status \cite{lombardo2022deepclasspathway}, a virus known to cause a variety of cancers (cite).   Overall, this method allows patient-specific identification of molecular pathways driving classifier decisions by mapping transcriptomic data to pathways. 

\paragraph{\textbf{Self-explainable neural networks}} 
DeepGONet, is a self-explainable deep fully-connected neural network, that is constrained by prior biological knowledge from GO \cite{bourgeais2021deep}.  In this model, each neuron has an associated GO function, therefore, predictions made by the network can be explained by the set of biological functions assigned to each neuron. DeepGONet is tasked with cancer-type classification and uses microarray and RNA-seq data as input.  Additional models have used similar techniques to identify cancer types such as PathDeep, which uses gene-to-pathway relationships from expression data and KEGG pathways \cite{park2021classification}.   

Cox-PASNet uses a self-explainable approach to integrate cancer gene expression data with clinical data for survival analysis. Nodes in the neural network correspond to genes and pathways, highlighting the nonlinear effects of these factors on cancer patient survival outcomes. This model architecture was chosen due to the highly non-linear and high-dimensional data with a low sample size that is typically seen in conventional survival analysis.  \cite{hao2019interpretable}. 

Fortelny et. al. have utilized knowledge-primed neural networks (KPNNs) on scRNA data showing their utility in identifying key regulators for cancer subtypes, and differentiation stages, and inferring cell identity.  Of note a new KPNN must be constructed for each task. In this network, each neuron has a molecular equivalent, such as a protein or gene, and every edge has a mechanistic interpretation, such as a regulatory interaction along a signaling pathway\cite{fortelny2020knowledge}.  

Liu et al. utilized similar methods to better understand important features of transcriptional control, in which a DNN was constructed with each layer having a specific chemical or biological interpretation \cite{liu2020fully}. When applied to the early embryo of fruit fly Drosophila the model was able to capture the chemistry of transcription factor binding to DNA.    

\paragraph{\textbf {Self-explainable autoencoders}} 

Models with an autoencoder base can also be supplemented with prior knowledge to increase interpretability.  Pathway module VAE (pmVAE) directly incorporates pathway information by restricting the structure of the VAE to mirror gene-pathway memberships. Thus, latent representations learned in this model can be factorized by pathway membership allowing for interpretable analysis of multiple downstream effects in scRNA-seq data, such as cell type and biological stimulus, within the contexts of each pathway \cite{gut2021pmvae}.  Similar methods have been used by Rybakov et al. using VAE with a regularized decoder and decomposing the scRNA-seq data into interpretable pathways, rather than using a factorization method \cite{rybakov2020learning}. 

Models that utilize variation within the dataset have additionally been leveraged to predict genetic/drug interactions. By first varying conditions such as drug dosage, time, drug type, and genetic knockouts, Lotfollahi et al. trained a compositional perturbation autoencoder (CPA) to predict expression patterns of interest in measure perturbation response across different combinations of conditions \cite{lotfollahi2021learning}. Each embedding in the CPA is separated and can be recombined to predict the effect of novel perturbation and covariate combinations. 

\paragraph{\textbf{Others}} 

DeepLIFT \cite{shrikumar2017learning} has been applied to DL approaches for predicting missing meta-data labels such as tissue, age, sex, etc. in small RNA expression profiles \cite{fiosina2020explainable}. Training on annotated human sRNA-seq samples from the sRNA expression atlas (SEA), the authors were able to identify relevant sRNAs for each prediction class, allowing for greater reusability of sequencing data.

\subsection{Bioimaging Informatics}

\begin{figure*}[!th]
    \centering
    \includegraphics[width=\textwidth]{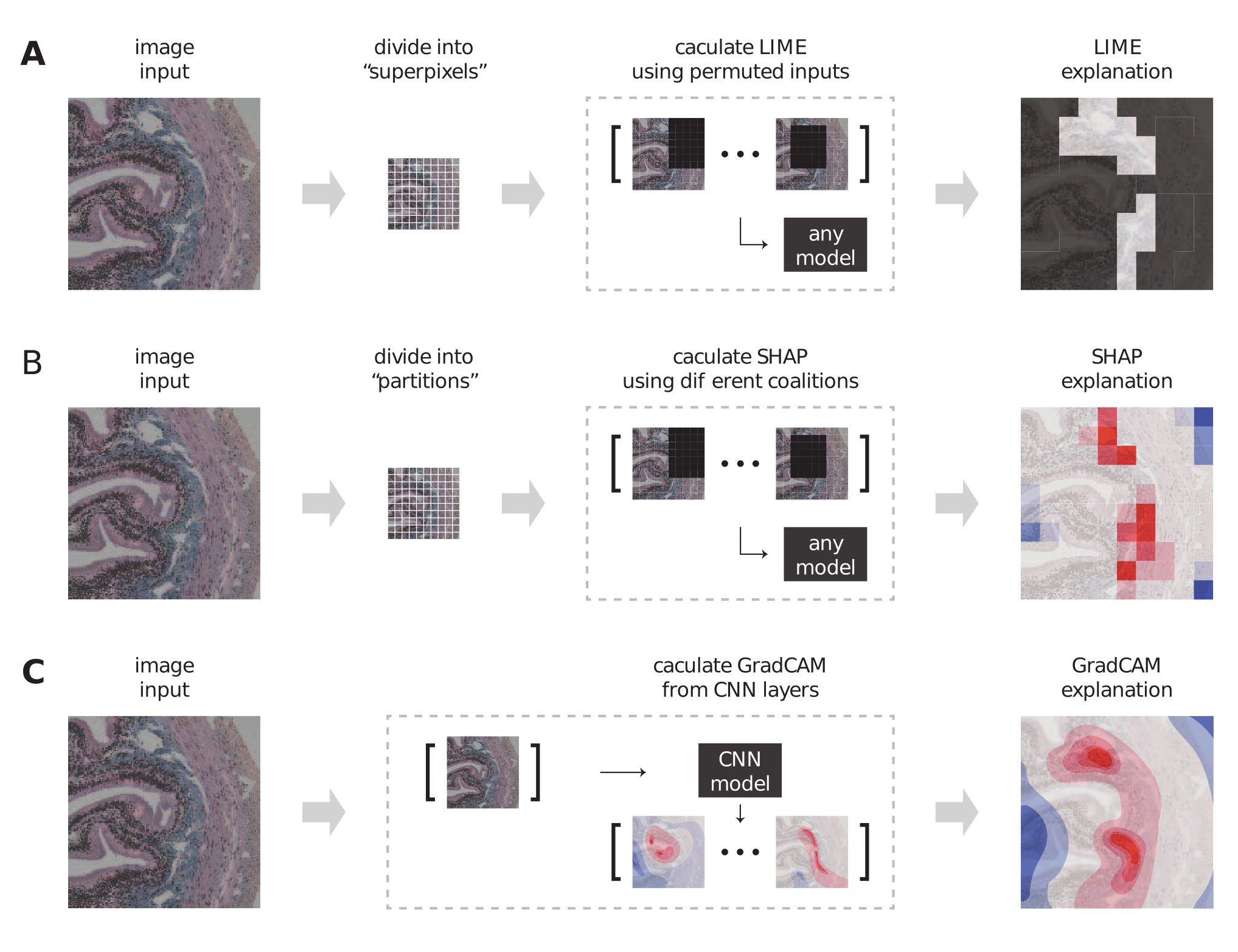}
    \vspace{-5mm}
    \caption{Bioimaging analysis uses CT or MRI images to create deep learning models. To explain these models, methods like dividing images into (A) "superpixels" (B) or "partitions" are employed, allowing XAI tools like LIME and SHAP to highlight crucial regions. (C) Additionally, Class Activation Maps generate heatmaps for CNN-based model explanations.}
    \label{fig:bioimage}
    \vspace{-2mm}
\end{figure*}

Explainable AI (XAI) techniques, including SHAP, LIME, and CAM, are increasingly employed in bioimaging informatics to elucidate model predictions. By generating heatmaps that highlight influential image regions, these methods enhance the reliability and trustworthiness of predictions from complex models.

\subsubsection{Model-agnostic Methods}

\paragraph{\textbf{SHAP}}

The SHapley Additive exPlanations (SHAP) method has been instrumental in indicating feature importance in trained models, enhancing the explainability of DL models \cite{severn2022pipeline}. For instance, SHAP has been used to clarify the results of a 3D regression CNN that assessed volumetric breast density, generating SHAP maps for each patient's breast MRI \cite{van2020volumetric}. The maps revealed that accurate and inaccurate estimations relied on different voxels. Specifically, accurate density estimations are mainly based on glandular and fatty tissue.

SHAP's ability to provide global and local explanations was harnessed in a recent study that predicts malignant cerebral edema after ischemic stroke using an LSTM model \cite{foroushani2022accelerating}. An image analysis workflow is developed to extract quantitative measurements from CT scans, which, along with clinical variables, were input features for the LSTM model. The global explanation of SHAP identified the hemispheric CSF ratio and the 24-hour NIHSS score as the overall significant predictors of edema, while age and ASPECTS contributed less. Moreover, individual SHAP values enabled personalized and interpretable predictions for malignant edema.

\paragraph{\textbf{LIME}} 

Local Interpretable Model-agnostic Explanations (LIME) employs ``superpixels'' as features for images, generated by dividing images into patches. Feature permutation involves enabling or disabling these superpixels by preserving or substituting pixel values \cite{magesh2020explainable}. LIME has been used to explain VGG-16 with transfer learning for Parkinson's disease classification on DaTSCAN SPECT brain images, localizing relevant regions and validating the model fidelity \cite{magesh2020explainable}. Additionally, LIME has been applied to explain COVID-19 diagnosis based on CT lung scans \cite{madhavi2022efficient} and prostate cancer classification from ultrasound and MRI images \cite{hassan2022prostate}, highlighting relevant regions corresponding to classification results.

\paragraph{\textbf{Others}}

In~\cite{mohagheghi2022developing}, the authors introduced an explainable deep correction method incorporating auxiliary knowledge, such as object shape and boundary profiles, to refine CNN model outputs for 3D medical segmentation tasks. They tested their approach on liver CT datasets, using a boundary validation (BV) model trained on a corrected previous slice, to identify invalid boundary points of initial results based on any CNN segmentation model, and a 2D Patch segmentation (PS) model, trained on boundary profiles, to correct invalid segmentations based on boundary profiles. This formed the Boundary correction loop (BCL), which generated the final 3D image when the loop reached the last slice.

\subsubsection{Model-specific Methods}

\paragraph{\textbf{Class Activation Maps}}
Gradient-weighted Class Activation Mapping (Grad-CAM) \cite{wang2020explainable} \cite{altan2022deepoct} \cite{singh2021covidscreen} \cite{yang2022deep} and its variants \cite{karim2020deepcovidexplainer} have been widely used to explain CNN based bioimaging prediction in many applications, such as optical
coherence tomography (OCT), chest radiography (CXR) images, etc.

Due to the COVID-19 outbreak, many studies proposed new models to detect pneumonia or COVID-19. These models often used chest x-ray (CXR) images and relied on methods like Grad-CAM and its variants to explain and validate the model predictions \cite{nedumkunnel2021explainable, singh2021covidscreen, ukwuoma2022hybrid, karim2020deepcovidexplainer}. These methods help identify key regions that the model focused on, which assists in choosing the best model. For instance, Ukwuoma et al.~\cite{ukwuoma2022hybrid} used Grad-CAM to explain the identification of pneumonia from chest X-rays using a hybrid DL framework. This framework combined a group of pretrained models to extract features and a transformer encoder to detect pneumonia. The authors compared the Grad-CAM heatmaps of the hybrid model, the ensemble part of the model and other pretrained deep-learning models, and they found that the hybrid model had stronger and clearer visualization results than others. Karimte et al. \cite{karim2020deepcovidexplainer} compared the important regions of confirmed COVID-19 CXR images identified by Grad-CAM, Grad-CAM++, and layerwise relevance propagation (LRP). They found that Grad-CAM++ highlighted more precise and accurate features than Grad-CAM and LRP. As a result, they used Grad-CAM++ to explain the predictions of their proposed ensemble CNN method for COVID-19 detection.

In addition to chest x-ray applications, many methods have been adopted to provide transparency in optical coherence tomography (OCT). Altan \cite{altan2022deepoct} used Grad-CAM heatmaps to validate the regions responsible for the diagnosis of diabetic macular edema (DME). Their study showed that the proposed CNN-based DeepOCT technology can identify small macular edemas, indicating that the proposed model was not restricted to detecting extensive macular pathology, but can also identify affected areas such as macular edema regions. Wang et al. \cite{wang2020explainable} used Grad-CAM to explain the most representative and confusing prediction of the VGG-19 neural network, which aims to classify OCT. Instead of examining all images, they only chose the correctly classified images with the highest probability and low probability, as well as the misclassified images that have a high probability of prediction. The explanation showed that for correctly classified images with the highest probability, the proposed model could easily identify correct regions with AMD biomarkers. Moreover, in the case of correctly classified images with low prediction confidence, the DL model identified the correct regions containing AMD biomarkers but was hindered by the presence of noisy pixels. However, for misclassified images with high prediction confidence, they had two observations: first, the DL model failed to focus on the correct region, resulting in an incorrect prediction; second, in cases where an image displayed multiple AMD biomarkers, the DL model selected the most prominent one as the predicted label. Vasquez et al. \cite{vasquez2022interactive} applied Grad-CAM in a different way. Instead of using it to assess the correctness of model, they integarted the heatmaps of Grad-CAM with expert annotated regions to train an interactive model to classify retinal disease.

Grad-CAM has also been adopted to explain the trained multiheaded attention transformer-based model for detecting Malaria Parasites from blood cell images \cite{islam2022explainable}. The heat maps revealed that the lesion regions were significantly redder than other areas of the image, suggesting that these areas were primarily responsible for the malaria parasites. In addition, Grad-CAM could be used to visualize the specific electrocardiography (ECG) leads and portions of ECG waves that contributed to the prediction of myocardial infarction \cite{jahmunah2022explainable}.

\section{Discussion and Future Direction}
The exponential increase in biological data from high-throughput experimental methods has led to the development of complex AI models to decipher hidden relationships within the data. However, increasing complexity comes at the cost of interpretability, necessitating the use of Explainable Artificial Intelligence (XAI) to demystify ``black-box'' models for biologists. In this review, we systematically examined recent advancements in machine learning and XAI in bioinformatics, covering key research areas such as sequence analysis, biological structures, gene expression and genomics data, and biomedical imaging. We find that despite significant improvements in model precision, current XAI methodologies, often not specifically tailored for bioinformatics, may use assumptions that do not hold true for biological data, limiting the utility of XAI in bioinformatics.

To overcome these challenges, we need XAI methods that are designed for bioinformatics. Where text-based input might stand alone, a biological sequence does not exist in a vacuum. Therefore, these methods should consider unique aspects of biological data, integrating expert knowledge and providing clear and accurate explanations that match the complexity of biological systems. Here, we outline the limitations of XAI in various bioinformatics research areas and suggest research directions for future investigations.

\paragraph{\textbf{Biological Sequence}}

The goal of XAI methods in DL models dealing with the biological sequence is to highlight biological motifs, individuals, or clusters of residues critical to a predicted function. Yet it is important to acknowledge that these XAI methods were not originally designed for biological sequences, only adapted to text-based sequence data. Thus, there are key limitations to current XAI applications that can be improved upon for more accurate explainability.

A limitation of the SHAP method is that it does not account for positional importance when applied to sequence text, because this method is intended to be used on tabular data. As for the attention score methods, since scores are highly dependent on the broader context of the sequence, it is challenging to determine if scores generated for individual residues are due to their independent importance or their interaction with other characters. Similarly, Grad-CAM, originally developed for dealing with continuous data in image classification, struggles with discrete data, which is the bulk of sequence text input. It additionally resolves local sequence context, but may not detect long-range residue interactions. These limitations represent a long-term unmet need for developing XAI models specific for biological sequence, but until then, those interested in XAI may consider combining XAI approaches to offset potential limitations or uncover more information about the underlying motivations of a DL model.

\paragraph{\textbf{Biological Structure}}
Biological structures can be represented in multiple ways, and understanding the right representation for the input data type for the downstream prediction task is important. XAI methods such as SHAP, LIME, Grad-CAM, and attention scores can uncover biases made by certain representations of biological structure, and provide additional confidence for the input parameters. With this in mind, the SHAP method being originally designed for tabular input data can have a hard time accounting for the spatial constraints of biological structures. Grad-CAM will do well in exploring 2d and 3D representations but fall short with 1D representations due to how sequences need to account for long-range interactions. As LIME relying on the model to be locally faithful, it will also struggle with 1D representations because DL models can learn that protein sequences are not locally faithful to protein structure (where small changes in sequence can lead to massive changes to the structure and vice versa). Finally, attention scores will be difficult to differentiate the individual importance of a feature from confounding importance with other features. 

\paragraph{\textbf{Gene Expression and Genome Analysis}}
As deep learning methods on sequencing data are applied in clinical research settings, an array of issues arises with data privacy, as well as, ethical and legal considerations, in which XAI could play a critical role in determining whether a model is fair and unbiased. These biases come from training datasets, as well as, biological and technical sources of variation during sequencing. Thus, it is crucial to standardize and benchmark these methods to enhance reproducibility and facilitate comparisons across studies, while also maintaining human-centered explainability. The outcome of XAI applications in these clinical settings should improve the confidence of predictive models for precision medicine and bridge the gap between bench and bedside.  

\paragraph{\textbf{Bioimage informatics}}
Both LIME and SHAP methods focus on providing explanations at the superpixel level, while Gradient-based models offer pixel-level explanations through upsampled weighted feature maps. However, the granularity of these methods may be too coarse for certain diseases, such as pneumonia, which require more fine-grained explanations that directly pinpoint the lesion area. Additionally, although these methods identify important regions for prediction, their reliability remains unclear since no confidence scores for those important features are provided, which is crucial when assisting physicians in decision-making. Finally,  existing bioimaging XAI often only showcases the most promising results. This underscores the need for metrics to validate the overall performance of these methods, as their collective efficacy remains unknown.

In conclusion, as the application of machine learning and deep learning methods in bioinformatics continues to expand, there is a growing need for Explainable AI techniques specifically designed to address the unique challenges within the field. By refining and developing XAI methods that consider the complexities of biological data, researchers can better understand and trust the models they use, leading to more accurate and reliable insights in various bioinformatics applications. With continued research and innovation, the integration of Explainable AI in bioinformatics promises to unlock new discoveries and drive advancements in our understanding of complex biological systems.








\bibliographystyle{unsrtnat}
\bibliography{reference}  

\begin{thebibliography}{90}
\providecommand{\natexlab}[1]{#1}
\providecommand{\url}[1]{\texttt{#1}}
\expandafter\ifx\csname urlstyle\endcsname\relax
  \providecommand{\doi}[1]{doi: #1}\else
  \providecommand{\doi}{doi: \begingroup \urlstyle{rm}\Url}\fi

\bibitem[LeCun et~al.(2015)LeCun, Bengio, and Hinton]{lecun2015deep}
Yann LeCun, Yoshua Bengio, and Geoffrey Hinton.
\newblock Deep learning.
\newblock \emph{nature}, 521\penalty0 (7553):\penalty0 436--444, 2015.

\bibitem[Krizhevsky et~al.(2017)Krizhevsky, Sutskever, and Hinton]{krizhevsky2017imagenet}
Alex Krizhevsky, Ilya Sutskever, and Geoffrey~E Hinton.
\newblock Imagenet classification with deep convolutional neural networks.
\newblock \emph{Communications of the ACM}, 60\penalty0 (6):\penalty0 84--90, 2017.

\bibitem[Zhang et~al.(2018)Zhang, Geiger, Pohjalainen, Mousa, Jin, and Schuller]{zhang2018deep}
Zixing Zhang, J{\"u}rgen Geiger, Jouni Pohjalainen, Amr El-Desoky Mousa, Wenyu Jin, and Bj{\"o}rn Schuller.
\newblock Deep learning for environmentally robust speech recognition: An overview of recent developments.
\newblock \emph{ACM Transactions on Intelligent Systems and Technology (TIST)}, 9\penalty0 (5):\penalty0 1--28, 2018.

\bibitem[Vaswani et~al.(2017)Vaswani, Shazeer, Parmar, Uszkoreit, Jones, Gomez, Kaiser, and Polosukhin]{vaswani2017attention}
Ashish Vaswani, Noam Shazeer, Niki Parmar, Jakob Uszkoreit, Llion Jones, Aidan~N Gomez, {\L}ukasz Kaiser, and Illia Polosukhin.
\newblock Attention is all you need.
\newblock \emph{Advances in neural information processing systems}, 30, 2017.

\bibitem[Pierson and Gashler(2017)]{pierson2017deep}
Harry~A Pierson and Michael~S Gashler.
\newblock Deep learning in robotics: a review of recent research.
\newblock \emph{Advanced Robotics}, 31\penalty0 (16):\penalty0 821--835, 2017.

\bibitem[YAN et~al.(2022)YAN, ZHANG, ZUO, ZHANG, WANG, and MAO]{yan2022attentionsplice}
Wenjing YAN, Baoyu ZHANG, Min ZUO, Qingchuan ZHANG, Hong WANG, and Da~MAO.
\newblock Attentionsplice: An interpretable multi-head self-attention based hybrid deep learning model in splice site prediction.
\newblock \emph{Chinese Journal of Electronics}, 31\penalty0 (5):\penalty0 870--887, 2022.

\bibitem[Zhou et~al.(2023)Zhou, Yeung, Gravel, Salcedo, Soleymani, Li, and Kannan]{zhou2023phosformer}
Zhongliang Zhou, Wayland Yeung, Nathan Gravel, Mariah Salcedo, Saber Soleymani, Sheng Li, and Natarajan Kannan.
\newblock Phosformer: an explainable transformer model for protein kinase-specific phosphorylation predictions.
\newblock \emph{Bioinformatics}, 39\penalty0 (2):\penalty0 btad046, 2023.

\bibitem[Taujale et~al.(2021)Taujale, Zhou, Yeung, Moremen, Li, and Kannan]{taujale2021mapping}
Rahil Taujale, Zhongliang Zhou, Wayland Yeung, Kelley~W Moremen, Sheng Li, and Natarajan Kannan.
\newblock Mapping the glycosyltransferase fold landscape using interpretable deep learning.
\newblock \emph{Nature communications}, 12\penalty0 (1):\penalty0 1--12, 2021.

\bibitem[Abbasi et~al.(2020)Abbasi, Razzaghi, Poso, Amanlou, Ghasemi, and Masoudi-Nejad]{abbasi2020deepcda}
Karim Abbasi, Parvin Razzaghi, Antti Poso, Massoud Amanlou, Jahan~B Ghasemi, and Ali Masoudi-Nejad.
\newblock Deepcda: deep cross-domain compound--protein affinity prediction through lstm and convolutional neural networks.
\newblock \emph{Bioinformatics}, 36\penalty0 (17):\penalty0 4633--4642, 2020.

\bibitem[Mieth et~al.(2021)Mieth, Rozier, Rodriguez, H{\"o}hne, G{\"o}rnitz, and M{\"u}ller]{mieth2021deepcombi}
Bettina Mieth, Alexandre Rozier, Juan~Antonio Rodriguez, Marina~MC H{\"o}hne, Nico G{\"o}rnitz, and Klaus-Robert M{\"u}ller.
\newblock Deepcombi: explainable artificial intelligence for the analysis and discovery in genome-wide association studies.
\newblock \emph{NAR genomics and bioinformatics}, 3\penalty0 (3):\penalty0 lqab065, 2021.

\bibitem[Chereda et~al.(2021)Chereda, Bleckmann, Menck, Perera-Bel, Stegmaier, Auer, Kramer, Leha, and Bei{\ss}barth]{chereda2021explaining}
Hryhorii Chereda, Annalen Bleckmann, Kerstin Menck, J{\'u}lia Perera-Bel, Philip Stegmaier, Florian Auer, Frank Kramer, Andreas Leha, and Tim Bei{\ss}barth.
\newblock Explaining decisions of graph convolutional neural networks: patient-specific molecular subnetworks responsible for metastasis prediction in breast cancer.
\newblock \emph{Genome medicine}, 13:\penalty0 1--16, 2021.

\bibitem[Mohagheghi and Foruzan(2022)]{mohagheghi2022developing}
Saeed Mohagheghi and Amir~Hossein Foruzan.
\newblock Developing an explainable deep learning boundary correction method by incorporating cascaded x-dim models to improve segmentation defects in liver ct images.
\newblock \emph{Computers in Biology and Medicine}, 140:\penalty0 105106, 2022.

\bibitem[Ukwuoma et~al.(2022)Ukwuoma, Qin, Heyat, Akhtar, Bamisile, Muaad, Addo, and Al-Antari]{ukwuoma2022hybrid}
Chiagoziem~C Ukwuoma, Zhiguang Qin, Md~Belal~Bin Heyat, Faijan Akhtar, Olusola Bamisile, Abdullah~Y Muaad, Daniel Addo, and Mugahed~A Al-Antari.
\newblock A hybrid explainable ensemble transformer encoder for pneumonia identification from chest x-ray images.
\newblock \emph{Journal of Advanced Research}, 2022.

\bibitem[Mei et~al.(2021)Mei, Desrosiers, and Frasnelli]{mei2021machine}
Jie Mei, Christian Desrosiers, and Johannes Frasnelli.
\newblock Machine learning for the diagnosis of parkinson's disease: a review of literature.
\newblock \emph{Frontiers in aging neuroscience}, 13:\penalty0 633752, 2021.

\bibitem[Bagherian et~al.(2021)Bagherian, Sabeti, Wang, Sartor, Nikolovska-Coleska, and Najarian]{bagherian2021machine}
Maryam Bagherian, Elyas Sabeti, Kai Wang, Maureen~A Sartor, Zaneta Nikolovska-Coleska, and Kayvan Najarian.
\newblock Machine learning approaches and databases for prediction of drug--target interaction: a survey paper.
\newblock \emph{Briefings in bioinformatics}, 22\penalty0 (1):\penalty0 247--269, 2021.

\bibitem[Jumper et~al.(2021)Jumper, Evans, Pritzel, Green, Figurnov, Ronneberger, Tunyasuvunakool, Bates, {\v{Z}}{\'\i}dek, Potapenko, et~al.]{jumper2021highly}
John Jumper, Richard Evans, Alexander Pritzel, Tim Green, Michael Figurnov, Olaf Ronneberger, Kathryn Tunyasuvunakool, Russ Bates, Augustin {\v{Z}}{\'\i}dek, Anna Potapenko, et~al.
\newblock Highly accurate protein structure prediction with alphafold.
\newblock \emph{Nature}, 596\penalty0 (7873):\penalty0 583--589, 2021.

\bibitem[Kim(2014)]{kim-2014-convolutional}
Yoon Kim.
\newblock Convolutional neural networks for sentence classification.
\newblock In \emph{Proceedings of the 2014 Conference on Empirical Methods in Natural Language Processing ({EMNLP})}, pages 1746--1751, Doha, Qatar, October 2014. Association for Computational Linguistics.
\newblock \doi{10.3115/v1/D14-1181}.
\newblock URL \url{https://aclanthology.org/D14-1181}.

\bibitem[Lin et~al.(2022)Lin, Akin, Rao, Hie, Zhu, Lu, dos Santos~Costa, Fazel-Zarandi, Sercu, Candido, et~al.]{lin2022language}
Zeming Lin, Halil Akin, Roshan Rao, Brian Hie, Zhongkai Zhu, Wenting Lu, Allan dos Santos~Costa, Maryam Fazel-Zarandi, Tom Sercu, Sal Candido, et~al.
\newblock Language models of protein sequences at the scale of evolution enable accurate structure prediction.
\newblock \emph{BioRxiv}, 2022.

\bibitem[Yeung et~al.(2023{\natexlab{a}})Yeung, Zhou, Mathew, Gravel, Taujale, O’Boyle, Salcedo, Venkat, Lanzilotta, Li, et~al.]{yeung2023tree}
Wayland Yeung, Zhongliang Zhou, Liju Mathew, Nathan Gravel, Rahil Taujale, Brady O’Boyle, Mariah Salcedo, Aarya Venkat, William Lanzilotta, Sheng Li, et~al.
\newblock Tree visualizations of protein sequence embedding space enable improved functional clustering of diverse protein superfamilies.
\newblock \emph{Briefings in Bioinformatics}, 24\penalty0 (1):\penalty0 bbac619, 2023{\natexlab{a}}.

\bibitem[Yeung et~al.(2023{\natexlab{b}})Yeung, Zhou, Li, and Kannan]{yeung2023alignment}
Wayland Yeung, Zhongliang Zhou, Sheng Li, and Natarajan Kannan.
\newblock Alignment-free estimation of sequence conservation for identifying functional sites using protein sequence embeddings.
\newblock \emph{Briefings in Bioinformatics}, 24\penalty0 (1):\penalty0 bbac599, 2023{\natexlab{b}}.

\bibitem[Madhavi and Supraja(2022)]{madhavi2022efficient}
M~Madhavi and P~Supraja.
\newblock Efficient explainable deep learning technique for covid-19 diagnosis based on computed tomography scan images of lungs.
\newblock In \emph{AIP Conference Proceedings}, volume 2385, page 050001. AIP Publishing LLC, 2022.

\bibitem[Magesh et~al.(2020)Magesh, Myloth, and Tom]{magesh2020explainable}
Pavan~Rajkumar Magesh, Richard~Delwin Myloth, and Rijo~Jackson Tom.
\newblock An explainable machine learning model for early detection of parkinson's disease using lime on datscan imagery.
\newblock \emph{Computers in Biology and Medicine}, 126:\penalty0 104041, 2020.

\bibitem[Hassan et~al.(2022)Hassan, Islam, Uddin, Ghoshal, Hassan, Huda, and Fortino]{hassan2022prostate}
Md~Rafiul Hassan, Md~Fakrul Islam, Md~Zia Uddin, Goutam Ghoshal, Mohammad~Mehedi Hassan, Shamsul Huda, and Giancarlo Fortino.
\newblock Prostate cancer classification from ultrasound and mri images using deep learning based explainable artificial intelligence.
\newblock \emph{Future Generation Computer Systems}, 127:\penalty0 462--472, 2022.

\bibitem[van~der Velden et~al.(2020)van~der Velden, Janse, Ragusi, Loo, and Gilhuijs]{van2020volumetric}
Bas~HM van~der Velden, Markus~HA Janse, Max~AA Ragusi, Claudette~E Loo, and Kenneth~GA Gilhuijs.
\newblock Volumetric breast density estimation on mri using explainable deep learning regression.
\newblock \emph{Scientific Reports}, 10\penalty0 (1):\penalty0 1--9, 2020.

\bibitem[Foroushani et~al.(2022)Foroushani, Hamzehloo, Kumar, Chen, Heitsch, Slowik, Strbian, Lee, Marcus, and Dhar]{foroushani2022accelerating}
Hossein~Mohammadian Foroushani, Ali Hamzehloo, Atul Kumar, Yasheng Chen, Laura Heitsch, Agnieszka Slowik, Daniel Strbian, Jin-Moo Lee, Daniel~S Marcus, and Rajat Dhar.
\newblock Accelerating prediction of malignant cerebral edema after ischemic stroke with automated image analysis and explainable neural networks.
\newblock \emph{Neurocritical Care}, 36\penalty0 (2):\penalty0 471--482, 2022.

\bibitem[Yap et~al.(2021)Yap, Johnston, Foley, MacDonald, Kondrashova, Tran, Nones, Koufariotis, Bean, Pearson, et~al.]{yap2021verifying}
Melvyn Yap, Rebecca~L Johnston, Helena Foley, Samual MacDonald, Olga Kondrashova, Khoa~A Tran, Katia Nones, Lambros~T Koufariotis, Cameron Bean, John~V Pearson, et~al.
\newblock Verifying explainability of a deep learning tissue classifier trained on rna-seq data.
\newblock \emph{Scientific reports}, 11\penalty0 (1):\penalty0 2641, 2021.

\bibitem[Yu et~al.(2021)Yu, Kossinna, Li, Liao, and Zhang]{yu2021explainable}
Yang Yu, Pathum Kossinna, Qing Li, Wenyuan Liao, and Qingrun Zhang.
\newblock Explainable autoencoder-based representation learning for gene expression data.
\newblock \emph{bioRxiv}, pages 2021--12, 2021.

\bibitem[Nelson et~al.(2022)Nelson, Riva, and Cvejic]{nelson2022smash}
Michael~E Nelson, Simone~G Riva, and Ana Cvejic.
\newblock Smash: a scalable, general marker gene identification framework for single-cell rna-sequencing.
\newblock \emph{BMC bioinformatics}, 23\penalty0 (1):\penalty0 328, 2022.

\bibitem[Kuruc et~al.(2022)Kuruc, Binder, and Hess]{kuruc2022stratified}
Fabrizio Kuruc, Harald Binder, and Moritz Hess.
\newblock Stratified neural networks in a time-to-event setting.
\newblock \emph{Briefings in Bioinformatics}, 23\penalty0 (1):\penalty0 bbab392, 2022.

\bibitem[Lemsara et~al.(2020)Lemsara, Ouadfel, and Fr{\"o}hlich]{lemsara2020pathme}
Amina Lemsara, Salima Ouadfel, and Holger Fr{\"o}hlich.
\newblock Pathme: pathway based multi-modal sparse autoencoders for clustering of patient-level multi-omics data.
\newblock \emph{BMC bioinformatics}, 21:\penalty0 1--20, 2020.

\bibitem[Withnell et~al.(2021)Withnell, Zhang, Sun, and Guo]{withnell2021xomivae}
Eloise Withnell, Xiaoyu Zhang, Kai Sun, and Yike Guo.
\newblock Xomivae: an interpretable deep learning model for cancer classification using high-dimensional omics data.
\newblock \emph{Briefings in Bioinformatics}, 22\penalty0 (6):\penalty0 bbab315, 2021.

\bibitem[Yang et~al.(2023)Yang, Yu, Wu, and Zhang]{yang2023cfa}
Tzu-Hsien Yang, Yu-Huai Yu, Sheng-Hang Wu, and Fang-Yuan Zhang.
\newblock Cfa: An explainable deep learning model for annotating the transcriptional roles of cis-regulatory modules based on epigenetic codes.
\newblock \emph{Computers in Biology and Medicine}, 152:\penalty0 106375, 2023.

\bibitem[Gu et~al.(2022)Gu, Zhang, Xu, Chen, Liu, Wu, Mo, Hu, Liu, and Luo]{gu2022protein}
Yuliang Gu, Xiangzhou Zhang, Anqi Xu, Weiqi Chen, Kang Liu, Lijuan Wu, Shenglong Mo, Yong Hu, Mei Liu, and Qichao Luo.
\newblock Protein--ligand binding affinity prediction with edge awareness and supervised attention.
\newblock \emph{iScience}, page 105892, 2022.

\bibitem[Wang et~al.(2020)Wang, Lucas, Furst, Fawzi, and Raicu]{wang2020explainable}
Yiyang Wang, Mirtha Lucas, Jacob Furst, Amani~A Fawzi, and Daniela Raicu.
\newblock Explainable deep learning for biomarker classification of oct images.
\newblock In \emph{2020 IEEE 20th International Conference on Bioinformatics and Bioengineering (BIBE)}, pages 204--210. IEEE, 2020.

\bibitem[Altan(2022)]{altan2022deepoct}
Gokhan Altan.
\newblock Deepoct: An explainable deep learning architecture to analyze macular edema on oct images.
\newblock \emph{Engineering Science and Technology, an International Journal}, 34:\penalty0 101091, 2022.

\bibitem[Singh et~al.(2021)Singh, Pandey, and Babu]{singh2021covidscreen}
Rajeev~Kumar Singh, Rohan Pandey, and Rishie~Nandhan Babu.
\newblock Covidscreen: explainable deep learning framework for differential diagnosis of covid-19 using chest x-rays.
\newblock \emph{Neural Computing and Applications}, 33\penalty0 (14):\penalty0 8871--8892, 2021.

\bibitem[Yang et~al.(2022{\natexlab{a}})Yang, Mei, and Piccialli]{yang2022deep}
Yuting Yang, Gang Mei, and Francesco Piccialli.
\newblock A deep learning approach considering image background for pneumonia identification using explainable ai (xai).
\newblock \emph{IEEE/ACM Transactions on Computational Biology and Bioinformatics}, 2022{\natexlab{a}}.

\bibitem[Karim et~al.(2020)Karim, D{\"o}hmen, Cochez, Beyan, Rebholz-Schuhmann, and Decker]{karim2020deepcovidexplainer}
Md~Rezaul Karim, Till D{\"o}hmen, Michael Cochez, Oya Beyan, Dietrich Rebholz-Schuhmann, and Stefan Decker.
\newblock Deepcovidexplainer: explainable covid-19 diagnosis from chest x-ray images.
\newblock In \emph{2020 IEEE International Conference on Bioinformatics and Biomedicine (BIBM)}, pages 1034--1037. IEEE, 2020.

\bibitem[Nedumkunnel et~al.(2021)Nedumkunnel, George, Sowmya, Rosh, and Mayya]{nedumkunnel2021explainable}
Ishan~Mathew Nedumkunnel, Linu~Elizabeth George, Kamath~S Sowmya, Neil~Abraham Rosh, and Veena Mayya.
\newblock Explainable deep neural models for covid-19 prediction from chest x-rays with region of interest visualization.
\newblock In \emph{2021 2nd International Conference on Secure Cyber Computing and Communications (ICSCCC)}, pages 96--101. IEEE, 2021.

\bibitem[Vasquez et~al.(2022)Vasquez, Shakya, Wang, Furst, Tchoua, and Raicu]{vasquez2022interactive}
Mariana Vasquez, Suhev Shakya, Ian Wang, Jacob Furst, Roselyne Tchoua, and Daniela Raicu.
\newblock Interactive deep learning for explainable retinal disease classification.
\newblock In \emph{Medical Imaging 2022: Image Processing}, volume 12032, pages 148--155. SPIE, 2022.

\bibitem[Islam et~al.(2022)Islam, Nahiduzzaman, Goni, Sayeed, Anower, Ahsan, and Haider]{islam2022explainable}
Md~Robiul Islam, Md~Nahiduzzaman, Md~Omaer~Faruq Goni, Abu Sayeed, Md~Shamim Anower, Mominul Ahsan, and Julfikar Haider.
\newblock Explainable transformer-based deep learning model for the detection of malaria parasites from blood cell images.
\newblock \emph{Sensors}, 22\penalty0 (12):\penalty0 4358, 2022.

\bibitem[Jahmunah et~al.(2022)Jahmunah, Ng, Tan, Oh, and Acharya]{jahmunah2022explainable}
V~Jahmunah, EYK Ng, Ru-San Tan, Shu~Lih Oh, and U~Rajendra Acharya.
\newblock Explainable detection of myocardial infarction using deep learning models with grad-cam technique on ecg signals.
\newblock \emph{Computers in Biology and Medicine}, 146:\penalty0 105550, 2022.

\bibitem[Karim et~al.(2019)Karim, Cochez, Beyan, Decker, and Lange]{karim2019onconetexplainer}
Md~Rezaul Karim, Michael Cochez, Oya Beyan, Stefan Decker, and Christoph Lange.
\newblock Onconetexplainer: explainable predictions of cancer types based on gene expression data.
\newblock In \emph{2019 IEEE 19th International Conference on Bioinformatics and Bioengineering (BIBE)}, pages 415--422. IEEE, 2019.

\bibitem[Lombardo et~al.(2022)Lombardo, Hess, Kurz, Riboldi, Marschner, Baumeister, Lauber, Pflugradt, Walch, Canis, et~al.]{lombardo2022deepclasspathway}
Elia Lombardo, Julia Hess, Christopher Kurz, Marco Riboldi, Sebastian Marschner, Philipp Baumeister, Kirsten Lauber, Ulrike Pflugradt, Axel Walch, Martin Canis, et~al.
\newblock Deepclasspathway: Molecular pathway aware classification using explainable deep learning.
\newblock \emph{European Journal of Cancer}, 176:\penalty0 41--49, 2022.

\bibitem[Chen et~al.(2021)Chen, Cheong, and Siu]{chen2021xdeep}
Jiarui Chen, Hong~Hin Cheong, and Shirley~WI Siu.
\newblock xdeep-acpep: deep learning method for anticancer peptide activity prediction based on convolutional neural network and multitask learning.
\newblock \emph{Journal of chemical information and modeling}, 61\penalty0 (8):\penalty0 3789--3803, 2021.

\bibitem[Monteiro et~al.(2022)Monteiro, Sim{\~o}es, {\'A}vila, Abbasi, Oliveira, and Arrais]{monteiro2022explainable}
Nelson~RC Monteiro, Carlos~JV Sim{\~o}es, Henrique~V {\'A}vila, Maryam Abbasi, Jos{\'e}~L Oliveira, and Joel~P Arrais.
\newblock Explainable deep drug--target representations for binding affinity prediction.
\newblock \emph{BMC bioinformatics}, 23\penalty0 (1):\penalty0 1--24, 2022.

\bibitem[Mukherjee et~al.(2021)Mukherjee, Su, and Rajan]{mukherjee2021deep}
Arpan Mukherjee, An~Su, and Krishna Rajan.
\newblock Deep learning model for identifying critical structural motifs in potential endocrine disruptors.
\newblock \emph{Journal of chemical information and modeling}, 61\penalty0 (5):\penalty0 2187--2197, 2021.

\bibitem[Yang et~al.(2022{\natexlab{b}})Yang, Zhong, Zhao, and Chen]{yang2022mgraphdta}
Ziduo Yang, Weihe Zhong, Lu~Zhao, and Calvin Yu-Chian Chen.
\newblock Mgraphdta: deep multiscale graph neural network for explainable drug--target binding affinity prediction.
\newblock \emph{Chemical science}, 13\penalty0 (3):\penalty0 816--833, 2022{\natexlab{b}}.

\bibitem[Danilevicz et~al.(2022)Danilevicz, Gill, Tay~Fernandez, Petereit, Upadhyaya, Batley, Bennamoun, Edwards, and Bayer]{danilevicz2022dnabert}
Monica~F Danilevicz, Mitchell Gill, Cassandria~G Tay~Fernandez, Jakob Petereit, Shriprabha~R Upadhyaya, Jacqueline Batley, Mohammed Bennamoun, David Edwards, and Philipp~E Bayer.
\newblock Dnabert-based explainable lncrna identification in plant genome assemblies.
\newblock \emph{bioRxiv}, pages 2022--02, 2022.

\bibitem[Avsec et~al.(2021)Avsec, Agarwal, Visentin, Ledsam, Grabska-Barwinska, Taylor, Assael, Jumper, Kohli, and Kelley]{avsec2021effective}
{\v{Z}}iga Avsec, Vikram Agarwal, Daniel Visentin, Joseph~R Ledsam, Agnieszka Grabska-Barwinska, Kyle~R Taylor, Yannis Assael, John Jumper, Pushmeet Kohli, and David~R Kelley.
\newblock Effective gene expression prediction from sequence by integrating long-range interactions.
\newblock \emph{Nature methods}, 18\penalty0 (10):\penalty0 1196--1203, 2021.

\bibitem[Hu et~al.(2019)Hu, Xiao, Zhang, Li, Shi, Jiang, Zhang, Zhang, and Zeng]{hu2019deephint}
Hailin Hu, An~Xiao, Sai Zhang, Yangyang Li, Xuanling Shi, Tao Jiang, Linqi Zhang, Lei Zhang, and Jianyang Zeng.
\newblock Deephint: understanding hiv-1 integration via deep learning with attention.
\newblock \emph{Bioinformatics}, 35\penalty0 (10):\penalty0 1660--1667, 2019.

\bibitem[Tian et~al.(2021)Tian, Zhou, Li, Tan, Cui, Xu, Wei, Zhu, Jin, Cao, et~al.]{tian2021deephpv}
Rui Tian, Ping Zhou, Mengyuan Li, Jinfeng Tan, Zifeng Cui, Wei Xu, Jingyue Wei, Jingjing Zhu, Zhuang Jin, Chen Cao, et~al.
\newblock Deephpv: a deep learning model to predict human papillomavirus integration sites.
\newblock \emph{Briefings in Bioinformatics}, 22\penalty0 (4):\penalty0 bbaa242, 2021.

\bibitem[Liang et~al.(2021)Liang, Cui, Wu, Yu, Tian, Xie, Jin, Fan, Xie, Huang, et~al.]{liang2021deepebv}
Jiuxing Liang, Zifeng Cui, Canbiao Wu, Yao Yu, Rui Tian, Hongxian Xie, Zhuang Jin, Weiwen Fan, Weiling Xie, Zhaoyue Huang, et~al.
\newblock Deepebv: a deep learning model to predict epstein--barr virus (ebv) integration sites.
\newblock \emph{Bioinformatics}, 37\penalty0 (20):\penalty0 3405--3411, 2021.

\bibitem[Mao et~al.(2017)Mao, Kostka, and Chikina]{mao2017modeling}
Weiguang Mao, Dennis Kostka, and Maria Chikina.
\newblock Modeling enhancer-promoter interactions with attention-based neural networks.
\newblock \emph{bioRxiv}, page 219667, 2017.

\bibitem[Karimi et~al.(2020)Karimi, Wu, Wang, and Shen]{karimi2020explainable}
Mostafa Karimi, Di~Wu, Zhangyang Wang, and Yang Shen.
\newblock Explainable deep relational networks for predicting compound--protein affinities and contacts.
\newblock \emph{Journal of chemical information and modeling}, 61\penalty0 (1):\penalty0 46--66, 2020.

\bibitem[Karimi et~al.(2019)Karimi, Wu, Wang, and Shen]{karimi2019deepaffinity}
Mostafa Karimi, Di~Wu, Zhangyang Wang, and Yang Shen.
\newblock Deepaffinity: interpretable deep learning of compound--protein affinity through unified recurrent and convolutional neural networks.
\newblock \emph{Bioinformatics}, 35\penalty0 (18):\penalty0 3329--3338, 2019.

\bibitem[Gao et~al.(2018)Gao, Fokoue, Luo, Iyengar, Dey, Zhang, et~al.]{gao2018interpretable}
Kyle~Yingkai Gao, Achille Fokoue, Heng Luo, Arun Iyengar, Sanjoy Dey, Ping Zhang, et~al.
\newblock Interpretable drug target prediction using deep neural representation.
\newblock In \emph{IJCAI}, volume 2018, pages 3371--3377, 2018.

\bibitem[Bourgeais et~al.(2021)Bourgeais, Zehraoui, Ben~Hamdoune, and Hanczar]{bourgeais2021deep}
Victoria Bourgeais, Farida Zehraoui, Mohamed Ben~Hamdoune, and Blaise Hanczar.
\newblock Deep gonet: self-explainable deep neural network based on gene ontology for phenotype prediction from gene expression data.
\newblock \emph{BMC bioinformatics}, 22\penalty0 (10):\penalty0 1--25, 2021.

\bibitem[Park et~al.(2021)Park, Huang, and Ahn]{park2021classification}
Sangick Park, Eunchong Huang, and Taejin Ahn.
\newblock Classification and functional analysis between cancer and normal tissues using explainable pathway deep learning through rna-sequencing gene expression.
\newblock \emph{International Journal of Molecular Sciences}, 22\penalty0 (21):\penalty0 11531, 2021.

\bibitem[Hao et~al.(2019)Hao, Kim, Mallavarapu, Oh, and Kang]{hao2019interpretable}
Jie Hao, Youngsoon Kim, Tejaswini Mallavarapu, Jung~Hun Oh, and Mingon Kang.
\newblock Interpretable deep neural network for cancer survival analysis by integrating genomic and clinical data.
\newblock \emph{BMC medical genomics}, 12\penalty0 (10):\penalty0 1--13, 2019.

\bibitem[Fortelny and Bock(2020)]{fortelny2020knowledge}
Nikolaus Fortelny and Christoph Bock.
\newblock Knowledge-primed neural networks enable biologically interpretable deep learning on single-cell sequencing data.
\newblock \emph{Genome biology}, 21\penalty0 (1):\penalty0 1--36, 2020.

\bibitem[Liu et~al.(2020)Liu, Barr, and Reinitz]{liu2020fully}
Yi~Liu, Kenneth Barr, and John Reinitz.
\newblock Fully interpretable deep learning model of transcriptional control.
\newblock \emph{Bioinformatics}, 36\penalty0 (Supplement\_1):\penalty0 i499--i507, 2020.

\bibitem[Gut et~al.(2021)Gut, Stark, R{\"a}tsch, and Davidson]{gut2021pmvae}
Gilles Gut, Stefan~G Stark, Gunnar R{\"a}tsch, and Natalie~R Davidson.
\newblock Pmvae: Learning interpretable single-cell representations with pathway modules.
\newblock \emph{bioRxiv}, pages 2021--01, 2021.

\bibitem[Rybakov et~al.(2020)Rybakov, Lotfollahi, Theis, and Wolf]{rybakov2020learning}
Sergei Rybakov, Mohammad Lotfollahi, Fabian~J Theis, and F~Alexander Wolf.
\newblock Learning interpretable latent autoencoder representations with annotations of feature sets.
\newblock \emph{bioRxiv}, pages 2020--12, 2020.

\bibitem[Lotfollahi et~al.(2021)Lotfollahi, Susmelj, De~Donno, Ji, Ibarra, Wolf, Yakubova, Theis, and Lopez-Paz]{lotfollahi2021learning}
Mohammad Lotfollahi, Anna~Klimovskaia Susmelj, Carlo De~Donno, Yuge Ji, Ignacio~L Ibarra, F~Alexander Wolf, Nafissa Yakubova, Fabian~J Theis, and David Lopez-Paz.
\newblock Learning interpretable cellular responses to complex perturbations in high-throughput screens.
\newblock \emph{BioRxiv}, pages 2021--04, 2021.

\bibitem[Ribeiro et~al.(2016)Ribeiro, Singh, and Guestrin]{ribeiro2016should}
Marco~Tulio Ribeiro, Sameer Singh, and Carlos Guestrin.
\newblock " why should i trust you?" explaining the predictions of any classifier.
\newblock In \emph{Proceedings of the 22nd ACM SIGKDD international conference on knowledge discovery and data mining}, pages 1135--1144, 2016.

\bibitem[Lundberg and Lee(2017)]{lundberg2017unified}
Scott~M Lundberg and Su-In Lee.
\newblock A unified approach to interpreting model predictions.
\newblock \emph{Advances in neural information processing systems}, 30, 2017.

\bibitem[Shapley et~al.(1953)]{shapley1953value}
Lloyd~S Shapley et~al.
\newblock A value for n-person games.
\newblock 1953.

\bibitem[Zhou et~al.(2016)Zhou, Khosla, Lapedriza, Oliva, and Torralba]{zhou2016learning}
Bolei Zhou, Aditya Khosla, Agata Lapedriza, Aude Oliva, and Antonio Torralba.
\newblock Learning deep features for discriminative localization.
\newblock In \emph{Proceedings of the IEEE conference on computer vision and pattern recognition}, pages 2921--2929, 2016.

\bibitem[Selvaraju et~al.(2017)Selvaraju, Cogswell, Das, Vedantam, Parikh, and Batra]{selvaraju2017grad}
Ramprasaath~R Selvaraju, Michael Cogswell, Abhishek Das, Ramakrishna Vedantam, Devi Parikh, and Dhruv Batra.
\newblock Grad-cam: Visual explanations from deep networks via gradient-based localization.
\newblock In \emph{Proceedings of the IEEE international conference on computer vision}, pages 618--626, 2017.

\bibitem[Schoenfelder and Fraser(2019)]{schoenfelder2019long}
Stefan Schoenfelder and Peter Fraser.
\newblock Long-range enhancer--promoter contacts in gene expression control.
\newblock \emph{Nature Reviews Genetics}, 20\penalty0 (8):\penalty0 437--455, 2019.

\bibitem[Berman et~al.(2002)Berman, Nibu, Pfeiffer, Tomancak, Celniker, Levine, Rubin, and Eisen]{berman2002exploiting}
Benjamin~P Berman, Yutaka Nibu, Barret~D Pfeiffer, Pavel Tomancak, Susan~E Celniker, Michael Levine, Gerald~M Rubin, and Michael~B Eisen.
\newblock Exploiting transcription factor binding site clustering to identify cis-regulatory modules involved in pattern formation in the drosophila genome.
\newblock \emph{Proceedings of the National Academy of Sciences}, 99\penalty0 (2):\penalty0 757--762, 2002.

\bibitem[Tyagi et~al.(2015)Tyagi, Tuknait, Anand, Gupta, Sharma, Mathur, Joshi, Singh, Gautam, and Raghava]{tyagi2015cancerppd}
Atul Tyagi, Abhishek Tuknait, Priya Anand, Sudheer Gupta, Minakshi Sharma, Deepika Mathur, Anshika Joshi, Sandeep Singh, Ankur Gautam, and Gajendra~PS Raghava.
\newblock Cancerppd: a database of anticancer peptides and proteins.
\newblock \emph{Nucleic acids research}, 43\penalty0 (D1):\penalty0 D837--D843, 2015.

\bibitem[Desaphy et~al.(2015)Desaphy, Bret, Rognan, and Kellenberger]{desaphy2015sc}
J{\'e}r{\'e}my Desaphy, Guillaume Bret, Didier Rognan, and Esther Kellenberger.
\newblock sc-pdb: a 3d-database of ligandable binding sites—10 years on.
\newblock \emph{Nucleic acids research}, 43\penalty0 (D1):\penalty0 D399--D404, 2015.

\bibitem[Breathnach and Chambon(1981)]{breathnach1981organization}
Richard Breathnach and Pierre Chambon.
\newblock Organization and expression of eucaryotic split genes coding for proteins.
\newblock \emph{Annual review of biochemistry}, 50\penalty0 (1):\penalty0 349--383, 1981.

\bibitem[Kulakovskiy et~al.(2013)Kulakovskiy, Medvedeva, Schaefer, Kasianov, Vorontsov, Bajic, and Makeev]{kulakovskiy2013hocomoco}
Ivan~V Kulakovskiy, Yulia~A Medvedeva, Ulf Schaefer, Artem~S Kasianov, Ilya~E Vorontsov, Vladimir~B Bajic, and Vsevolod~J Makeev.
\newblock Hocomoco: a comprehensive collection of human transcription factor binding sites models.
\newblock \emph{Nucleic acids research}, 41\penalty0 (D1):\penalty0 D195--D202, 2013.

\bibitem[Liu et~al.(2022)Liu, Cheng, Ashraf, Zhang, Wang, Lv, He, Song, and Zuo]{liu2022interpretation}
Shang Liu, Hailiang Cheng, Javaria Ashraf, Youping Zhang, Qiaolian Wang, Limin Lv, Man He, Guoli Song, and Dongyun Zuo.
\newblock Interpretation of convolutional neural networks reveals crucial sequence features involving in transcription during fiber development.
\newblock \emph{BMC bioinformatics}, 23\penalty0 (1):\penalty0 91, 2022.

\bibitem[Khan et~al.(2018)Khan, Fornes, Stigliani, Gheorghe, Castro-Mondragon, Van Der~Lee, Bessy, Cheneby, Kulkarni, Tan, et~al.]{khan2018jaspar}
Aziz Khan, Oriol Fornes, Arnaud Stigliani, Marius Gheorghe, Jaime~A Castro-Mondragon, Robin Van Der~Lee, Adrien Bessy, Jeanne Cheneby, Shubhada~R Kulkarni, Ge~Tan, et~al.
\newblock Jaspar 2018: update of the open-access database of transcription factor binding profiles and its web framework.
\newblock \emph{Nucleic acids research}, 46\penalty0 (D1):\penalty0 D260--D266, 2018.

\bibitem[Kurumbail et~al.(1996)Kurumbail, Stevens, Gierse, McDonald, Stegeman, Pak, Gildehaus, Iyashiro, Penning, Seibert, et~al.]{kurumbail1996structural}
Ravi~G Kurumbail, Anna~M Stevens, James~K Gierse, Joseph~J McDonald, Roderick~A Stegeman, Jina~Y Pak, Daniel Gildehaus, Julie~M Iyashiro, Thomas~D Penning, Karen Seibert, et~al.
\newblock Structural basis for selective inhibition of cyclooxygenase-2 by anti-inflammatory agents.
\newblock \emph{Nature}, 384\penalty0 (6610):\penalty0 644--648, 1996.

\bibitem[Wang et~al.(2013)Wang, Jiang, Ma, Wu, Wacker, Katritch, Han, Liu, Huang, Vardy, et~al.]{wang2013structural}
Chong Wang, Yi~Jiang, Jinming Ma, Huixian Wu, Daniel Wacker, Vsevolod Katritch, Gye~Won Han, Wei Liu, Xi-Ping Huang, Eyal Vardy, et~al.
\newblock Structural basis for molecular recognition at serotonin receptors.
\newblock \emph{Science}, 340\penalty0 (6132):\penalty0 610--614, 2013.

\bibitem[Luo et~al.(2017)Luo, Chen, Cheng, Ma, Li, Zhang, Li, Zhang, Guo, Li, et~al.]{luo2017allosteric}
Qichao Luo, Liping Chen, Xi~Cheng, Yuqin Ma, Xiaona Li, Bing Zhang, Li~Li, Shilei Zhang, Fei Guo, Yang Li, et~al.
\newblock An allosteric ligand-binding site in the extracellular cap of k2p channels.
\newblock \emph{Nature communications}, 8\penalty0 (1):\penalty0 378, 2017.

\bibitem[Rodr{\'\i}guez et~al.(2014)Rodr{\'\i}guez, Brea, Loza, and Carlsson]{rodriguez2014structure}
David Rodr{\'\i}guez, Jos{\'e} Brea, Mar{\'\i}a~Isabel Loza, and Jens Carlsson.
\newblock Structure-based discovery of selective serotonin 5-ht1b receptor ligands.
\newblock \emph{Structure}, 22\penalty0 (8):\penalty0 1140--1151, 2014.

\bibitem[Ehrlich et~al.(2017)Ehrlich, G{\"o}ller, and Grimme]{ehrlich2017towards}
Stephan Ehrlich, Andreas~H G{\"o}ller, and Stefan Grimme.
\newblock Towards full quantum-mechanics-based protein--ligand binding affinities.
\newblock \emph{ChemPhysChem}, 18\penalty0 (8):\penalty0 898--905, 2017.

\bibitem[Alves et~al.(2016)Alves, Muratov, Capuzzi, Politi, Low, Braga, Zakharov, Sedykh, Mokshyna, Farag, et~al.]{alves2016alarms}
Vinicius~M Alves, Eugene~N Muratov, Stephen~J Capuzzi, Regina Politi, Yen Low, Rodolpho~C Braga, Alexey~V Zakharov, Alexander Sedykh, Elena Mokshyna, Sherif Farag, et~al.
\newblock Alarms about structural alerts.
\newblock \emph{Green Chemistry}, 18\penalty0 (16):\penalty0 4348--4360, 2016.

\bibitem[Casals-Casas and Desvergne(2011)]{casals2011endocrine}
Cristina Casals-Casas and B{\'e}atrice Desvergne.
\newblock Endocrine disruptors: from endocrine to metabolic disruption.
\newblock \emph{Annual review of physiology}, 73:\penalty0 135--162, 2011.

\bibitem[Bung et~al.(2022)Bung, Krishnan, and Roy]{bung2022silico}
Navneet Bung, Sowmya~Ramaswamy Krishnan, and Arijit Roy.
\newblock An in silico explainable multiparameter optimization approach for de novo drug design against proteins from the central nervous system.
\newblock \emph{Journal of Chemical Information and Modeling}, 62\penalty0 (11):\penalty0 2685--2695, 2022.

\bibitem[Kavlock et~al.(2012)Kavlock, Chandler, Houck, Hunter, Judson, Kleinstreuer, Knudsen, Martin, Padilla, Reif, et~al.]{kavlock2012update}
Robert Kavlock, Kelly Chandler, Keith Houck, Sid Hunter, Richard Judson, Nicole Kleinstreuer, Thomas Knudsen, Matt Martin, Stephanie Padilla, David Reif, et~al.
\newblock Update on epa’s toxcast program: providing high throughput decision support tools for chemical risk management.
\newblock \emph{Chemical research in toxicology}, 25\penalty0 (7):\penalty0 1287--1302, 2012.

\bibitem[Shrikumar et~al.(2017)Shrikumar, Greenside, and Kundaje]{shrikumar2017learning}
Avanti Shrikumar, Peyton Greenside, and Anshul Kundaje.
\newblock Learning important features through propagating activation differences.
\newblock In \emph{International conference on machine learning}, pages 3145--3153. PMLR, 2017.

\bibitem[Fiosina et~al.(2020)Fiosina, Fiosins, and Bonn]{fiosina2020explainable}
Jelena Fiosina, Maksims Fiosins, and Stefan Bonn.
\newblock Explainable deep learning for augmentation of small rna expression profiles.
\newblock \emph{Journal of Computational Biology}, 27\penalty0 (2):\penalty0 234--247, 2020.

\bibitem[Severn et~al.(2022)Severn, Suresh, G{\"o}rg, Choi, Jain, and Ghosh]{severn2022pipeline}
Cameron Severn, Krithika Suresh, Carsten G{\"o}rg, Yoon~Seong Choi, Rajan Jain, and Debashis Ghosh.
\newblock A pipeline for the implementation and visualization of explainable machine learning for medical imaging using radiomics features.
\newblock \emph{Sensors}, 22\penalty0 (14):\penalty0 5205, 2022.

\end{thebibliography}

\end{document}